%% file: main.tex
\documentclass[12pt]{article}
 \usepackage{graphicx}
 \usepackage{amssymb}
 \usepackage{amsmath}
 \usepackage{amsbsy}
 \usepackage{latexsym}
 \usepackage{subfigure}
\usepackage{amsthm}
\usepackage{multirow}
\usepackage{epsfig} % for postscript graphics files
\usepackage{pseudocode}
\usepackage{url}

 \input{mathdef.tex}

\input{mathdef_suzuki.tex}
\usepackage{setspace} %{doublespace}

 \usepackage{mlapa}
\begin{document}

\title{\vspace*{-33mm}
\begin{flushleft}
  \normalsize
  \sl
% Submitted to
% % %To appear in 
% \textit{Neural Computation}.
% % %  vol.xxx, no.xxx, pp.xxx-xxx, 2013.
\end{flushleft}
\vspace*{5mm}
Density-Difference Estimation
}

\author{
Masashi Sugiyama\\
Tokyo Institute of Technology, Japan.\\
sugi@cs.titech.ac.jp\\
http://sugiyama-www.cs.titech.ac.jp/\~{}sugi\\[2mm]
Takafumi Kanamori\\
Nagoya University, Japan.\\
kanamori@is.nagoya-u.ac.jp\\[2mm]
Taiji Suzuki\\
The University of Tokyo, Japan.\\
s-taiji@stat.t.u-tokyo.ac.jp\\[2mm]
Marthinus Christoffel du Plessis\\
Tokyo Institute of Technology, Japan.\\
christo@sg.cs.titech.ac.jp\\[2mm]
Song Liu\\
Tokyo Institute of Technology, Japan.\\
song@sg.cs.titech.ac.jp\\[2mm]
Ichiro Takeuchi\\
Nagoya Institute of Technology, Japan.\\
takeuchi.ichiro@nitech.ac.jp\\
}

\date{}

 \sloppy
\maketitle
\thispagestyle{myheadings}
\markright{}

  \begin{abstract}\noindent
We address the problem of estimating the \emph{difference} between
two probability densities.
A naive approach 
is a two-step procedure of first estimating two densities separately
and then computing their difference.
However, such a two-step procedure does not necessarily work well
because the first step is performed without regard to the second step
and thus a small error incurred in the first stage can cause a big error in the second stage.
In this paper, we propose a single-shot procedure 
for directly estimating the density difference
without separately estimating two densities.
We derive a non-parametric finite-sample error bound
for the proposed single-shot density-difference estimator
and show that it achieves the optimal convergence rate.
The usefulness of the proposed method is also demonstrated experimentally.
% We apply the proposed method to $L^2$-distance estimation and show that 
% it is more accurate than the difference of kernel density estimators
% and it is more robust against outliers than
% Kullback-Leibler divergence estimation.

\begin{center}
  \textbf{Keywords}
\end{center}
density difference,
$L^2$-distance,
robustness,
Kullback-Leibler divergence,
kernel density estimation.
%quadratic mutual information.
% Bregman divergence.

\end{abstract}

\pagestyle{myheadings}
\markright{Density-Difference Estimation}

\doublespacing

%%%%%%%%%%%%%%%%%%%%%%%%%%%%%%%%%%%%%%%%%%%%%%%%%%%%%%%%%%%%
\section{Introduction}\label{sec:introduction}
When estimating a quantity consisting of two elements,
a two-stage approach of first
estimating the two elements separately and then approximating
the target quantity based on the estimates of the two elements
often performs poorly,
because the first stage is carried out without regard to the second stage
and thus a small error incurred in the first stage can cause a big error in the second stage.
To cope with this problem, it would be more appropriate
to directly estimate the target quantity in a single-shot process
without separately estimating the two elements.

A seminal example that follows this general idea is pattern recognition
by the \emph{support vector machine}
\cite{COLT:Boser+etal:1992,mach:Cortes+Vapnik:1995,book:Vapnik:1998}:
Instead of separately estimating two probability distributions of patterns
for positive and negative classes, 
the support vector machine directly learns the boundary between the two classes
that is sufficient for pattern recognition.
More recently, a problem of estimating the ratio of two probability densities
was tackled in a similar fashion
\cite{Biometrika:Qin:1998,AISM:Sugiyama+etal:2008,Inbook:Gretton+etal:2009,JMLR:Kanamori+etal:2009,IEEE-IT:Nguyen+etal:2010,ML:Kanamori+etal:2012,AISM:Sugiyama+etal:2012,book:Sugiyama+etal:2012}:
The ratio of two probability densities is directly estimated without going through
separate estimation of the two probability densities.

In this paper, we further explore this line of research,
and propose a method for directly estimating the \emph{difference}
between two probability densities in a single-shot process.
Density differences are useful for various purposes such as 
class-balance estimation under class-prior change
\cite{nc:Saerens+Latinne+Decaestecker:2002,ICML:duPlessis+Sugiyama:2012},
change-point detection in time series \cite{SAM:Kawahara+Sugiyama:2012,arXiv:Song+etal:2012},
feature extraction \cite{JMLR:Torkkola:2003},
video-based event detection \cite{VECTaR:Yamanaka+etal:2011},
flow cytometric data analysis \cite{BJ:Duong+etal:2009},
ultrasound image segmentation \cite{PR:Liu+eta:2010},
non-rigid image registration \cite{ANBC:Atif+etal:2003},
and
image-based target recognition \cite{SPIE:Gray+Principe:2010}.

For this density-difference estimation problem,
we propose a single-shot method, called
the \emph{least-squares density-difference} (LSDD) estimator,
that directly estimates the density difference without
separately estimating two densities.
LSDD is derived within a framework of kernel least-squares estimation,
and its solution can be computed \emph{analytically} in a computationally efficient
and stable manner.
Furthermore, LSDD is equipped with cross-validation,
and thus all tuning parameters such as the kernel width
and the regularization parameter can be systematically and objectively optimized.
We derive a finite-sample error bound for the LSDD estimator
in a non-parametric setup
and show that it achieves the optimal convergence rate.
% Moreover, we theoretically show that LSDD in the non-parametric setup
% is superior to the two-stage procedure
% by the difference of kernel density estimators (KDEs)
% in terms of the convergence rate.

We also apply LSDD to $L^2$-distance estimation and show that it is more accurate than
the difference of KDEs,
which tends to severely under-estimate the $L^2$-distance \cite{JMA:Anderson+etal:1994}.
Compared with the 
\emph{Kullback-Leibler (KL) divergence} \cite{Annals-Math-Stat:Kullback+Leibler:1951},
the $L^2$-distance is more robust against outliers
\cite{Biometrika:Basu+etal:1998,Tech:Scott:2001,CSDA:Besbeas+Morgan:2004}.

Finally, we experimentally demonstrate the usefulness of LSDD
in semi-supervised class-prior estimation
and
unsupervised change detection.

The rest of this paper is structured as follows.
In Section~\ref{sec:LSDD}, we derive the LSDD method
and investigate its theoretical properties.
In Section~\ref{sec:L2-distance},
we show how the $L^2$-distance can be approximated by LSDD.
In Section~\ref{sec:experiments}, we illustrate 
the numerical behavior of LSDD.
% We further show possible variations of LSDD in Section~\ref{sec:variations},
Finally, we conclude in Section~\ref{sec:conclusion}.

%%%%%%%%%%%%%%%%%%%%%%%%%%%%%%%%%%%%%%%%%%%%%%%%%%%%%%%%%%%%
\section{Density-Difference Estimation}
\label{sec:LSDD}
In this section, we propose a single-shot method for estimating
the difference between two probability densities from samples,
and analyze its theoretical properties.

\subsection{Problem Formulation and Naive Approach}\label{subsec:DKDE}
First, we formulate the problem of density-difference estimation.

Suppose that we are given two sets of independent and identically distributed samples
$\calX:=\{\boldx_i\}_{i=1}^{\nsample}$ and $\calX':=\{\boldx'_{i'}\}_{i'=1}^{\nsample'}$
drawn from probability distributions on $\mathbbR^\inputdim$
with densities $\density(\boldx)$ and $\density'(\boldx)$, respectively:
\begin{align*}
\calX&:=\{\boldx_i\}_{i=1}^{\nsample}\iid \density(\boldx),\\
\calX'&:=\{\boldx'_{i'}\}_{i'=1}^{\nsample'}\iid \density'(\boldx).
\end{align*}
Our goal is to estimate the difference
$\diff(\boldx)$ between $\density(\boldx)$ and $\density'(\boldx)$
from the samples $\calX$ and $\calX'$:
\begin{align*}
  \diff(\boldx):=\density(\boldx)-\density'(\boldx).
\end{align*}

% $L^2$-distance between
% $\density(\boldx)$ and $\density'(\boldx)$,
% \begin{align}
%   L^2(\density,\density'):=\int\left(\density(\boldx)-\density'(\boldx)\right)^2\mathrm{d}\boldx,
%   \label{L2dist}
% \end{align}
% from the samples $\calX$ and $\calX'$.

A naive approach to density-difference estimation is
to use \emph{kernel density estimators} (KDEs)
\cite{book:Silverman:1986}.
For Gaussian kernels, the KDE-based density-difference estimator is given by
\begin{align*}
 \widetilde{\diff}(\boldx):=
\densityh(\boldx)-\densityh'(\boldx),
\end{align*}
where
\begin{align*}
\densityh(\boldx)&:=\frac{1}{\nsample(2\pi\sigma^2)^{\inputdim/2}}\sum_{i=1}^{\nsample}
  \exp\left(-\frac{\|\boldx-\boldx_i\|^2}{2\sigma^2}\right),\\
\densityh'(\boldx)&:=\frac{1}{\nsample'(2\pi\sigma'^2)^{\inputdim/2}}\sum_{i'=1}^{\nsample'}
  \exp\left(-\frac{\|\boldx-\boldx'_{i'}\|^2}{2\sigma'^2}\right).
\end{align*}
The Gaussian widths $\sigma$ and $\sigma'$ may be determined
based on cross-validation \cite{book:Haerdle+etal:2004}.

However, we argue that the KDE-based density-difference estimator is not 
the best approach because of its two-step nature:
Small estimation error in each density estimate
can cause a big error in the final density-difference estimate.
More intuitively, good density estimators tend to be smooth
and thus a density-difference estimator
obtained from such smooth density estimators tends to be over-smoothed
\cite[see also numerical experiments in Section~\ref{subsec:experiments-LSDDvsKDE}]{Biometrika:Hall+Wand:1988,JMA:Anderson+etal:1994}.

To overcome this weakness, we give a single-shot procedure of
directly estimating the density difference $f(\boldx)$ without separately estimating
the densities $\density(\boldx)$ and $\density'(\boldx)$.

\subsection{Least-Squares Density-Difference Estimation}
\label{subsec:LSDD}
% The $L^2$-distance \eqref{L2dist} can expressed as
% \begin{align}
%   L^2(\density,\density')&=\int\diff(\boldx)^2\mathrm{d}\boldx,
%   \label{L2dist2}
% \end{align}
% where 

% Our basic idea is to directly estimate the density difference $\diff(\boldx)$
% without estimating the densities $\density(\boldx)$ and $\density'(\boldx)$.

% More specifically, 
In our proposed approach, we fit a density-difference model
$\diffmodel(\boldx)$ 
%(where $\boldtheta$ is a parameter)
to the true density-difference function $\diff(\boldx)$
under the squared loss:
% \footnote{
% \emcite{Biometrika:Hall+Wand:1988} used a leave-one-out variant of
% this criterion for jointly determining the bandwidths of two KDEs.
% }:
\begin{align}
%  \boldtheta^\ast:=\argmin_{\boldtheta}
  \argmin_{\diffmodel}
  \int\left(\diffmodel(\boldx)-\diff(\boldx)\right)^2
  \mathrm{d}\boldx.
  \label{boldtheta-ast}
\end{align}
We use the following linear-in-parameter model as $\diffmodel(\boldx)$:
\begin{align}
  \diffmodel(\boldx)=\sum_{\ell=1}^{\nparam}\theta_\ell\psi_\ell(\boldx)
  =\boldtheta^\top\boldpsi(\boldx), 
  \label{linear-model}
\end{align}
where $\nparam$ denotes the number of basis functions,
$\boldpsi(\boldx)=(\psi_1(\boldx),\ldots,\psi_{\nparam}(\boldx))^{\top}$ is
a $\nparam$-dimensional basis function vector,
$\boldtheta=(\theta_1,\ldots,\theta_{\nparam})^\top$
is a $\nparam$-dimensional parameter vector,
and $^\top$ denotes the transpose.
In practice, we use the following non-parametric Gaussian kernel model as $\diffmodel(\boldx)$:
\begin{align}
  \diffmodel(\boldx)
  =\sum_{\ell=1}^{\nsample+\nsample'}\theta_\ell
  \exp\left(-\frac{\|\boldx-\boldc_\ell\|^2}{2\sigma^2}\right),
  \label{Gaussian-kernel-model}
\end{align}
where
$
  (\boldc_1,\ldots,\boldc_{\nsample},\boldc_{\nsample+1},\ldots,\boldc_{\nsample+\nsample'})
  :=(\boldx_1,\ldots,\boldx_{\nsample},\boldx'_1,\ldots,\boldx'_{\nsample'})
$
are Gaussian kernel centers.
If $\nsample+\nsample'$ is large,
we may use only a subset of
$\boldx_1,\ldots,\boldx_{\nsample},\boldx'_1,\ldots,\boldx'_{\nsample'}$
as Gaussian kernel centers.

For the model \eqref{linear-model},
the optimal parameter $\boldtheta^\ast$ is given by
\begin{align*}
  \boldtheta^\ast&:=\argmin_{\boldtheta}
  \int\left(\diffmodel(\boldx)-\diff(\boldx)\right)^2
  \mathrm{d}\boldx\\
  &\phantom{:}=\argmin_{\boldtheta}
  \left[
    \int\diffmodel(\boldx)^2\mathrm{d}\boldx
    -2\int\diffmodel(\boldx)\diff(\boldx)\mathrm{d}\boldx
  \right]\\
  &\phantom{:}=\argmin_{\boldtheta}
  \left[
    \boldtheta^\top\boldH\boldtheta-2\boldh^\top\boldtheta
  \right]\\
  &\phantom{:}=\boldH^{-1}\boldh,
\end{align*}
where $\boldH$ is the $\nparam\times\nparam$ matrix
and $\boldh$ is the $\nparam$-dimensional vector defined as
\begin{align*}
 \boldH&:=\int\boldpsi(\boldx)\boldpsi(\boldx)^{\top}\mathrm{d}\boldx,\\
 \boldh&:=\int\boldpsi(\boldx)\density(\boldx)\mathrm{d}\boldx
-\int\boldpsi(\boldx')\density'(\boldx')\mathrm{d}\boldx'.
% \boldE_{\density}[\boldpsi]-\boldE_{\density'}[\boldpsi].
\end{align*}
% $\boldE_\density[\boldpsi]$ denotes the expectation of
% $\boldpsi(\boldx)$ under the probability density $\density(\boldx)$:
% \begin{align*}
%   \boldE_\density[\boldpsi]:=\int\boldpsi(\boldx)\density(\boldx)\mathrm{d}\boldx.
% \end{align*}
% \begin{align*}
%   J_0(\boldtheta)=
%   \int\diffmodel(\boldx)^2\mathrm{d}\boldx
%   -2\int\diffmodel(\boldx)\diff(\boldx)\mathrm{d}\boldx+C,
% \end{align*}
% where $C$ is a constant independent of $\boldtheta$.
% We denote $J_0$ without $C$ by $J$:
% \begin{align*}
%   J(\boldtheta)&:=J_0(\boldtheta)-C\\
%   &\phantom{:}=\int\diffmodel(\boldx)^2\mathrm{d}\boldx
%   -2\int\diffmodel(\boldx)\diff(\boldx)\mathrm{d}\boldx\\
%   &\phantom{:}=\boldtheta^\top\boldH\boldtheta
%   -2\boldtheta^\top\boldh,
% \end{align*}
Note that, for the Gaussian kernel model \eqref{Gaussian-kernel-model},
the integral in $\boldH$ can be computed analytically as
\begin{align*}
  H_{\ell,\ell'}&=\int\exp\left(-\frac{\|\boldx-\boldc_\ell\|^2}{2\sigma^2}\right)
  \exp\left(-\frac{\|\boldx-\boldc_{\ell'}\|^2}{2\sigma^2}\right)\mathrm{d}\boldx\\
  &=(\pi\sigma^2)^{\inputdim/2}
  \exp\left(-\frac{\|\boldc_{\ell}-\boldc_{\ell'}\|^2}{4\sigma^2}\right),
%   h_{\ell}&=\int\exp\left(-\frac{\|\boldx-\boldc_\ell\|^2}{2\sigma^2}\right)
%   \density(\boldx)\mathrm{d}\boldx
%   -\int\exp\left(-\frac{\|\boldx-\boldc_\ell\|^2}{2\sigma^2}\right)
%   \density'(\boldx)\mathrm{d}\boldx,
\end{align*}
where $\inputdim$ denotes the dimensionality of $\boldx$.

Replacing the expectations in $\boldh$ by empirical estimators
and adding an $\ell_2$-regularizer to the objective function,
we arrive at the following optimization problem:
\begin{align}
\boldthetah:=\argmin_{\boldtheta}
\left[\boldtheta^\top\boldH\boldtheta-2\boldhh^\top\boldtheta+\lambda\boldtheta^\top\boldtheta\right],
\label{LSDD}
\end{align}
where $\lambda$ ($\ge0$) is the regularization parameter
and $\boldhh$ is the $\nparam$-dimensional vector defined as
% \begin{align*}
%   \boldhh&:=\widehat{\boldE}_{\density}[\boldpsi]-\widehat{\boldE}_{\density'}[\boldpsi].
% \end{align*}
% $\widehat{\boldE}_\density[\boldpsi]$ is an empirical estimator
% of $\boldE_\density[\boldpsi]$ given, e.g., by
\begin{align*}
  \boldhh&=\frac{1}{\nsample}\sum_{i=1}^{\nsample}\boldpsi(\boldx_i)
  -\frac{1}{\nsample'}\sum_{i'=1}^{\nsample'}\boldpsi(\boldx'_{i'}).
\end{align*}
% \begin{align*}
%   \hh_{\ell}&:=\frac{1}{\nsample}\sum_{i=1}^{\nsample}
%   \exp\left(-\frac{\|\boldx_i-\boldc_\ell\|^2}{2\sigma^2}\right)
%   -\frac{1}{\nsample'}\sum_{i'=1}^{\nsample'}
%   \exp\left(-\frac{\|\boldx'_{i'}-\boldc_\ell\|^2}{2\sigma^2}\right).
% \end{align*}
Taking the derivative of the objective function in Eq.\eqref{LSDD}
and equating it to zero,
we can obtain the solution $\boldthetah$ analytically as
\begin{align*}
  \boldthetah=\left(\boldH+\lambda\boldI_{\nparam}\right)^{-1}\boldhh,
\end{align*}
where $\boldI_{\nparam}$ denotes 
the $\nparam$-dimensional identity matrix.

Finally, a density-difference estimator $\diffh(\boldx)$ is given as
\begin{align}
  \diffh(\boldx)
  =\boldthetah^\top\boldpsi(\boldx).
\label{LSDD-fhat}
\end{align}
% \begin{align}
%   \diffh(\boldx)
%   =\sum_{\ell=1}^{\nparam}\thetah_\ell
%   \exp\left(-\frac{\|\boldx-\boldc_\ell\|^2}{2\sigma^2}\right).
% \label{LSDD-fhat}
% \end{align}
We call this the \emph{least-squares density-difference} (LSDD) estimator.

\subsection{Theoretical Analysis}
\label{sec:theory}
Here, we theoretically investigate the behavior of the LSDD estimator.

\subsubsection{Parametric Convergence}
First, we consider a linear parametric setup where
basis functions in our density-difference model \eqref{linear-model} are fixed.

Suppose that $\nsample/(\nsample+\nsample')$ converges to $\eta\in[0,1]$.
Then the \emph{central limit theorem} \cite{book:Rao:1965} asserts that
$\sqrt{\frac{\nsample\nsample'}{\nsample+\nsample'}}(\boldthetah-\boldtheta^\ast)$
converges in law to the normal distribution
with mean $\boldzero$ and covariance matrix
\begin{align*}
\boldH^{-1}((1-\eta)\boldV_\density+\eta\boldV_{\density'})\boldH^{-1},  
\end{align*}
where $\boldV_\density$ denotes the covariance matrix of 
$\boldpsi(\boldx)$ under the probability density $\density(\boldx)$:
% \begin{align*}
%   \boldV_\density[\boldpsi]:=\boldE_\density\left[
%   \left(\boldpsi-\boldE_\density[\boldpsi]\right)
%   \left(\boldpsi-\boldE_\density[\boldpsi]\right)^\top
%   \right].
% \end{align*}
\begin{align}
  \boldV_\density:=\int
  \left(\boldpsi(\boldx)-\boldpsi_\density\right)
  \left(\boldpsi(\boldx)-\boldpsi_\density\right)^\top
  \density(\boldx)\mathrm{d}\boldx,
\label{boldV}
\end{align}
and $\boldpsi_\density$ denotes the expectation of
$\boldpsi(\boldx)$ under the probability density $\density(\boldx)$:
\begin{align*}
  \boldpsi_\density:=\int\boldpsi(\boldx)\density(\boldx)\mathrm{d}\boldx.
\end{align*}

This result implies that the LSDD estimator has asymptotic normality
with asymptotic order $\sqrt{1/\nsample+1/\nsample'}$,
which is the optimal convergence rate in the parametric setup.

\subsubsection{Non-Parametric Error Bound}
\label{subsec:nonparametric-convergence}
Next, we consider a non-parametric setup
where a density-difference function is learned 
in a Gaussian \emph{reproducing kernel Hilbert space} (RKHS) \cite{AMS:Aronszajn:1950}.

Let $\calHg$ be the Gaussian RKHS with width $\gamma$:
\[
k_{\gamma}(\boldx,\boldx') = \exp\left(- \frac{\|\boldx - \boldx'\|^2}{\gamma^2} \right).
\]
Let us consider a slightly modified LSDD estimator 
that is more suitable for non-parametric error analysis:
For $\nsample'=\nsample$,
\begin{align*}
\diffh := \mathop{\arg \min }_{g \in \calHg}\left[
\|g\|_{L^2}^2 
-2\left(\frac{1}{\nsample}\sum_{i=1}^{\nsample} g(\boldx_i)
+\frac{1}{\nsample}\sum_{i'=1}^{\nsample} g(\boldx'_{i'})\right)
 + \lambda \|g\|^2_{\calHg}
\right],
\end{align*}
where $\|\cdot\|_{L^2}$ denotes the $L^2$-norm
and $\|\cdot\|_{\calHg}$ denotes the norm in RKHS $\calHg$.

Then we can prove that, for all $\rho,\rho'>0$,
there exists a constant $K$ such that,
for all $\tau\ge1$ and $n\ge1$,
the non-parametric LSDD estimator with
appropriate choice of $\lambda$ and $\gamma$ satisfies\footnote{
Because our theoretical result is highly technical,
we only describe a rough idea here.
More precise statement of the result
and its complete proof are provided in Appendix~\ref{proof-th:TheMainBound},
where we utilize the mathematical technique developed in \emcite{NIPS2011_0874}
for a regression problem.}
\begin{align}
\|\diffh - \diff \|_{L^2}^2 + \lambda \|\diffh\|_{\calHg}^2 \leq  
K \left( n^{-\frac{2\alpha}{2\alpha + d} + \rho} + \tau n^{-1+\rho'} \right),
\end{align}
with probability not less than $1- 4e^{-\tau}$.
Here, $d$ denotes the dimensionality of input vector $\boldx$, and
$\alpha\ge0$ denotes the regularity of Besov space to which 
the true density-difference function $\diff$ belongs
(smaller/larger $\alpha$ means $\diff$ is ``less/more complex'';
see Appendix~\ref{proof-th:TheMainBound} for its precise definition).
Because $n^{-\frac{2\alpha}{2\alpha + d}}$ is the optimal learning rate in this setup
\cite{NIPS2011_0874},
the above result shows that
the non-parametric LSDD estimator achieves the optimal convergence rate.

It is known that, if the naive KDE with a Gaussian kernel
is used for estimating a probability density with regularity $\alpha > 2$, 
the optimal learning rate cannot be achieved \cite{AMS:Farrell:1972,book:Silverman:1986}.
To achieve the optimal rate by KDE, we should choose a kernel
specifically tailored to each regularity $\alpha$ \cite{AMS:Parzen:1962}.
But such a kernel is not non-negative and it is difficult to implement in practice.
On the other hand, our LSDD estimator can always achieve the optimal learning rate
with a Gaussian kernel without regard to regularity $\alpha$.

\subsection{Model Selection by Cross-Validation}
\label{sec:LSDD-CV}
The above theoretical analyses showed the superiority of LSDD.
However, the practical performance of LSDD depends on the choice of models
(i.e., the kernel width $\sigma$ and the regularization parameter $\lambda$).
Here, we show that the model can be optimized by \emph{cross-validation} (CV).

More specifically, we first divide the samples
$\calX=\{\boldx_i\}_{i=1}^{\nsample}$
and $\calX'=\{\boldx'_{i'}\}_{i'=1}^{\nsample'}$
into $T$ disjoint subsets
$\{\calX_t\}_{t=1}^{T}$ and $\{\calX'_t\}_{t=1}^{T}$, respectively.
Then we obtain a density-difference estimate $\diffh_t(\boldx)$
from $\calX\backslash\calX_t$ and $\calX'\backslash\calX'_t$
(i.e., all samples without $\calX_t$ and $\calX'_t$),
and compute its hold-out error for $\calX_t$ and $\calX'_t$ as
\begin{align*}
  \mathrm{CV}^{(t)}
  &:=
    \int\!\diffh_t(\boldx)^2\mathrm{d}\boldx
    -\frac{2}{|\calX_t|}\!\sum_{\boldx\in\calX_t}\!\!\diffh_t(\boldx)
    +\frac{2}{|\calX'_t|}\!\sum_{\boldx'\in\calX'_t}\!\!\diffh_t(\boldx'),
\end{align*}
where $|\calX|$ denotes the number of elements in the set $\calX$.
We repeat this hold-out validation procedure for $t=1,\ldots,T$,
and compute the average hold-out error as
\begin{align*}
  \mathrm{CV}
  &:=
  \frac{1}{T}\sum_{t=1}^T \mathrm{CV}^{(t)}.
\end{align*}
Finally, we choose the model that minimizes $\mathrm{CV}$.

A MATLAB$^\text{\textregistered}$ implementation of
LSDD is available from
\begin{center}
\url{http://sugiyama-www.cs.titech.ac.jp/~sugi/software/LSDD/}'.
\\
(to be made public after acceptance)
\end{center}

%%%%%%%%%%%%%%%%%%%%%%%%%%%%%%%%%%%%%%%%%%%%%%%%%%%%%%%%%%%%
\section{$L^2$-Distance Estimation by LSDD}\label{sec:L2-distance}
In this section, 
we consider the problem of approximating the $L^2$-distance between
$\density(\boldx)$ and $\density'(\boldx)$,
\begin{align}
  L^2(\density,\density'):=\int\left(\density(\boldx)-\density'(\boldx)\right)^2\mathrm{d}\boldx,
  \label{L2-distance}
\end{align}
from samples
$\calX:=\{\boldx_i\}_{i=1}^{\nsample}$ and $\calX':=\{\boldx'_{i'}\}_{i'=1}^{\nsample'}$
(see Section~\ref{subsec:DKDE}).
% Our proposed estimator is given by
% \begin{align*}
%   \widehat{L}^2(\density,\density'):=
%   2\boldthetah^\top\boldhh-\boldthetah^\top\boldH\boldthetah
%   -\tr{\boldH^{-1}\left(\frac{1}{\nsample}\widehat{\boldV}_{\density}[\boldpsi]
%       -\frac{1}{\nsample'}\widehat{\boldV}_{\density'}[\boldpsi]\right)},
% \end{align*}

\subsection{Basic Form}
For an equivalent expression
\begin{align*}
  L^2(\density,\density')&=
  \int f(\boldx)\density(\boldx)\mathrm{d}\boldx
  -\int f(\boldx')\density'(\boldx')\mathrm{d}\boldx',
\end{align*}
if we replace $\diff(\boldx)$ with an LSDD estimator $\diffh(\boldx)$
and approximate the expectations by empirical averages,
the following $L^2$-distance estimator can be obtained:
\begin{align}
  L^2(\density,\density')\approx\boldhh^\top\boldthetah.
  \label{L2h-theta-hh}
\end{align}
Similarly, for another expression
\begin{align*}
L^2(\density,\density')
&=\int\diff(\boldx)^2\mathrm{d}\boldx,
\end{align*}
replacing $\diff(\boldx)$ with an LSDD estimator $\diffh(\boldx)$
gives another $L^2$-distance estimator:
\begin{align}
  L^2(\density,\density')\approx\boldthetah^\top\boldH\boldthetah.
  \label{L2h-theta-H-theta}
\end{align}

\subsection{Reduction of Bias Caused by Regularization}
Eq.\eqref{L2h-theta-hh} and Eq.\eqref{L2h-theta-H-theta}
themselves give approximations to $L^2(\density,\density')$.
Nevertheless, we argue that the use of their combination, defined by
\begin{align}
  \widehat{L}^2(\calX,\calX'):=
  2\boldhh^\top\boldthetah-\boldthetah^\top\boldH\boldthetah,
  \label{L2h-2theta-hh--theta-H-theta}
\end{align}
is more sensible. 
To explain the reason,
let us consider a generalized $L^2$-distance estimator of the following form:
\begin{align}
  \beta\boldhh^\top\boldthetah+(1-\beta)\boldthetah^\top\boldH\boldthetah,
  \label{L2h-alpha-theta-hh--(1-alpha)-theta-H-theta}
\end{align}
where $\beta$ is a real scalar.
If the regularization parameter $\lambda$ ($\ge0$) is small,
then Eq.\eqref{L2h-alpha-theta-hh--(1-alpha)-theta-H-theta} can be expressed as
\begin{align}
  \beta\boldhh^\top\boldthetah+(1-\beta)\boldthetah^\top\boldH\boldthetah
  =\boldhh^\top\boldH^{-1}\boldhh-\lambda(2-\beta)\boldhh^\top\boldH^{-2}\boldhh
  +o_p(\lambda),
  \label{L2h-alpha-theta-hh--(1-alpha)-theta-H-theta-expansion}
\end{align}
where $o_p$ denotes the probabilistic order
(its derivation is given in Appendix~\ref{proof-theo:L2h-alpha-theta-hh--(1-alpha)-theta-H-theta-expansion}).

Thus, the bias introduced by regularization 
(i.e., the second term in the right-hand side of 
Eq.\eqref{L2h-alpha-theta-hh--(1-alpha)-theta-H-theta-expansion}
that depends on $\lambda$)
can be eliminated if $\beta=2$, which yields Eq.\eqref{L2h-2theta-hh--theta-H-theta}.
Note that, if no regularization is imposed (i.e., $\lambda=0$),
both Eq.\eqref{L2h-theta-hh} and Eq.\eqref{L2h-theta-H-theta}
yield $\boldhh^\top\boldH^{-1}\boldhh$,
the first term in the right-hand side of
Eq.\eqref{L2h-alpha-theta-hh--(1-alpha)-theta-H-theta-expansion}.

Eq.\eqref{L2h-2theta-hh--theta-H-theta}
is actually equivalent to the negative of the optimal objective value 
of the LSDD optimization problem without regularization (i.e., Eq.\eqref{LSDD} with $\lambda=0$).
This can be naturally interpreted through
a lower bound of $L^2(\density,\density')$ 
obtained by \emph{Legendre-Fenchel convex duality} \cite{book:Rockafellar:1970}:
\begin{align*}
  L^2(\density,\density')=\sup_{g}
  \left[
    2\left(
      \int g(\boldx)\density(\boldx)\mathrm{d}\boldx
      -\int g(\boldx)\density'(\boldx)\mathrm{d}\boldx
    \right)
    -\int g(\boldx)^2\mathrm{d}\boldx
  \right],
\end{align*}
where the supremum is attained at $g=\diff$.
If the expectations are replaced by empirical estimators
and the linear-in-parameter model \eqref{linear-model} is used as $g$,
the above optimization problem is reduced
to the LSDD objective function without regularization (see Eq.\eqref{LSDD}).
Thus, LSDD corresponds to approximately maximizing the above lower bound
and Eq.\eqref{L2h-2theta-hh--theta-H-theta} is its maximum value.

Through eigenvalue decomposition of $\boldH$, we can show that
\begin{align*}
  2\boldhh^\top\boldthetah-\boldthetah^\top\boldH\boldthetah
  \ge
  \boldhh^\top\boldthetah
  \ge
  \boldthetah^\top\boldH\boldthetah.
\end{align*}
Thus, our approximator \eqref{L2h-2theta-hh--theta-H-theta}
is not less than the plain approximators
\eqref{L2h-theta-hh} and \eqref{L2h-theta-H-theta}.

%we use our LSDD estimator $\diffh(\boldx)$ as $g(\boldx)$
%and replace the expectations by empirical estimators,
% we obtain Eq.\eqref{L2h-2theta-hh--theta-H-theta}.
% Thus, LSDD corresponds to maximizing the above variational lower bound.

\subsection{Further Bias Correction}
$\boldhh^\top\boldH^{-1}\boldhh$,
the first term in Eq.\eqref{L2h-alpha-theta-hh--(1-alpha)-theta-H-theta-expansion},
is an essential part of the $L^2$-distance estimator
\eqref{L2h-2theta-hh--theta-H-theta}.
However, it is actually a slightly biased estimator
of the target quantity $\boldh^\top\boldH^{-1}\boldh$
($=\boldtheta^*{}^\top\boldH\boldtheta^*=\boldh^\top\boldtheta^*$):
\begin{align}  
\mathbbE[\boldhh^\top\boldH^{-1}\boldhh]
=\boldh^\top\boldH^{-1}\boldh
+\tr{\boldH^{-1}\left(\frac{1}{\nsample}\boldV_{\density}
    +\frac{1}{\nsample'}\boldV_{\density'}\right)},
\label{hh-Hinv-hh-bias}
\end{align}
where $\mathbbE$ denotes the expectation over all samples
$\calX=\{\boldx_i\}_{i=1}^{\nsample}$ and $\calX'=\{\boldx'_{i'}\}_{i'=1}^{\nsample'}$,
and $\boldV_{\density}$ and $\boldV_{\density'}$ are defined by Eq.\eqref{boldV}
(its derivation is given in Appendix~\ref{proof-theo:hh-Hinv-hh-bias}).

The second term in the right-hand side of Eq.\eqref{hh-Hinv-hh-bias}
is an estimation bias that is generally non-zero.
Thus, based on Eq.\eqref{hh-Hinv-hh-bias}, we can construct a bias-corrected
$L^2$-distance estimator as
\begin{align}
\widetilde{L}^2(\calX,\calX'):=
2\boldhh^\top\boldthetah-\boldthetah^\top\boldH\boldthetah
  -\tr{\boldH^{-1}\left(\frac{1}{\nsample}\widehat{\boldV}_{\density}
      +\frac{1}{\nsample'}\widehat{\boldV}_{\density'}\right)},
   \label{L2h-bias-correction}
\end{align}
where $\widehat{\boldV}_{\density}$ is an empirical estimator
of covariance matrix $\boldV_{\density}$:
\begin{align*}
  \widehat{\boldV}_\density:=
  \frac{1}{\nsample}\sum_{i=1}^{\nsample}
  \left(\boldpsi(\boldx_i)-\widehat{\boldpsi}_\density\right)
  \left(\boldpsi(\boldx_i)-\widehat{\boldpsi}_\density\right)^\top,
\end{align*}
and $\widehat{\boldpsi}_\density$ is an empirical estimator
of the expectation $\boldpsi_\density$:
\begin{align*}
  \widehat{\boldpsi}_\density:=\frac{1}{\nsample}\sum_{i=1}^{\nsample}\boldpsi(\boldx_i).
\end{align*}

% \subsection{Positive-Part Estimator}
The true $L^2$-distance is non-negative by definition
(see Eq.\eqref{L2-distance}),
% However, given that the bias-correction term
% $\tr{\boldH^{-1}\left(\frac{1}{\nsample}\widehat{\boldV}_{\density}[\boldpsi]
%       +\frac{1}{\nsample'}\widehat{\boldV}_{\density'}[\boldpsi]\right)}$
% in Eq.\eqref{L2h-bias-correction} is positive, 
but the above bias-corrected estimate can take a negative value.
Following the same line as \emcite{TechRepo:Baranchik:1964},
the \emph{positive-part} estimator may be more accurate:
\begin{align*}
  \overline{L}^2(\calX,\calX'):=
  \max\left\{0,\widetilde{L}^2(\calX,\calX')\right\}.
\end{align*}
However, in our preliminary experiments,
$\overline{L}^2(\calX,\calX')$ does not always perform well 
particularly when $\boldH$ is ill-conditioned.
For this reason, we practically propose to use $\widehat{L}^2(\calX,\calX')$
defined by Eq.\eqref{L2h-2theta-hh--theta-H-theta}.

\section{Experiments}
\label{sec:experiments}
In this section, we experimentally evaluate the performance of LSDD.

\subsection{Numerical Examples}
First, we show numerical examples using artificial datasets.

\begin{figure*}[p]
  \centering
  \begin{minipage}[t]{0.40\textwidth}
  \centering
    \includegraphics[width=1\textwidth,clip]{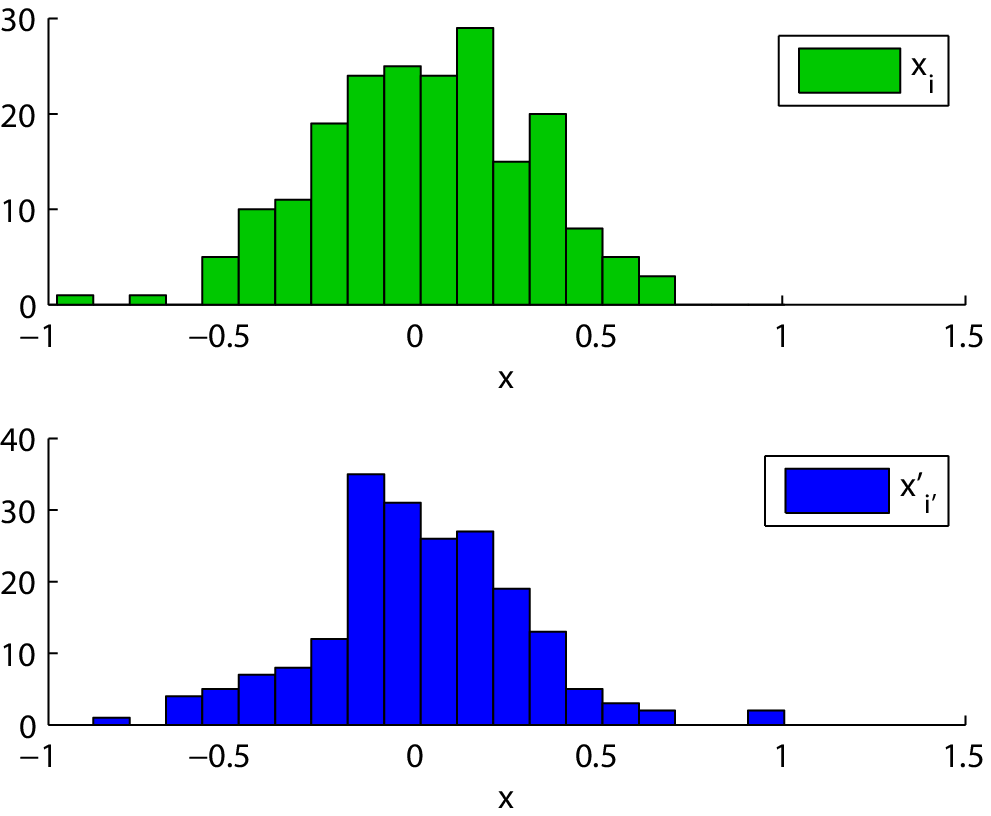}\\
    (a) Samples\\[4mm]
    \includegraphics[width=1\textwidth,clip]{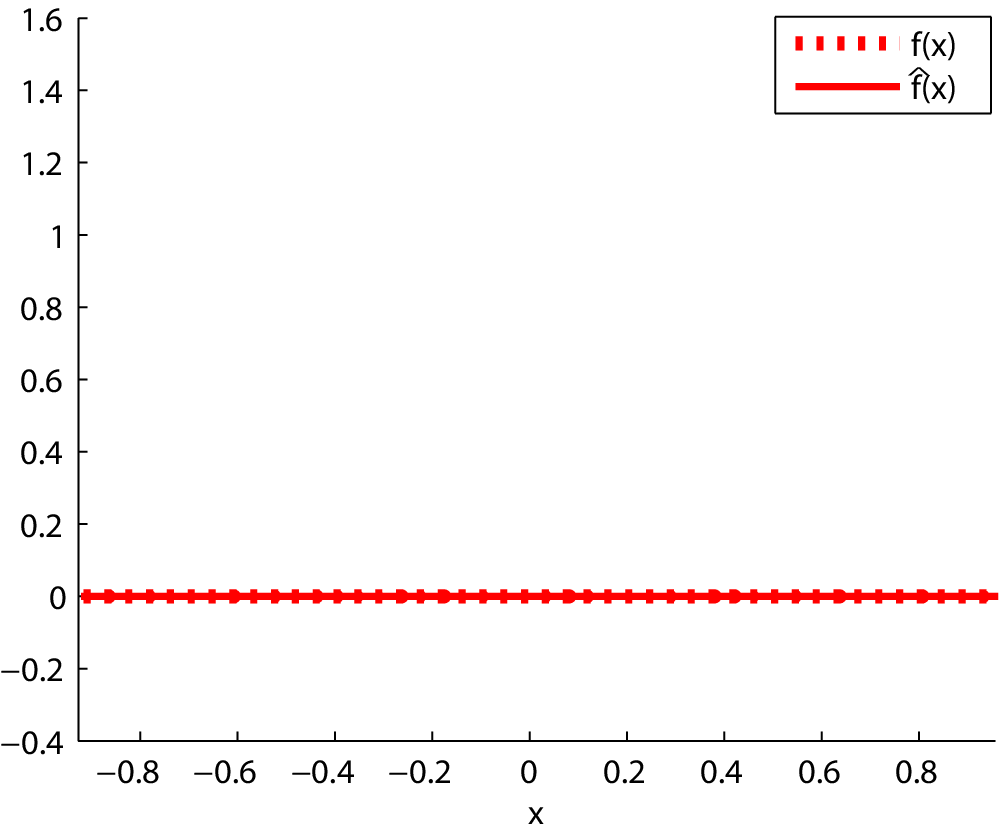}\\
    (b) LSDD\\[4mm]
    \includegraphics[width=1\textwidth,clip]{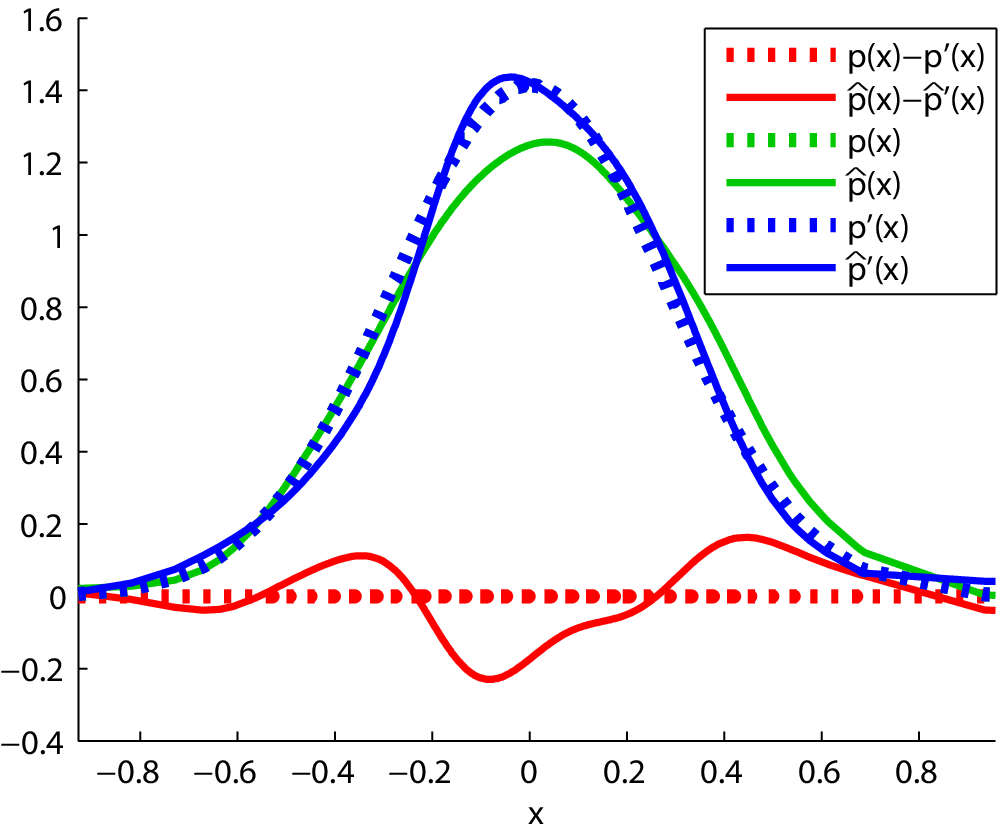}\\
    (c) KDE
  \caption{
    Estimation of density difference when $\mu=0$ (i.e., $f(x)=\density(x)-\density'(x)=0$).
  }
  \label{fig:illustration-DD1-n200}
  \end{minipage}
~~~~~~~~~~
  \begin{minipage}[t]{0.40\textwidth}
  \centering
    \includegraphics[width=1\textwidth,clip]{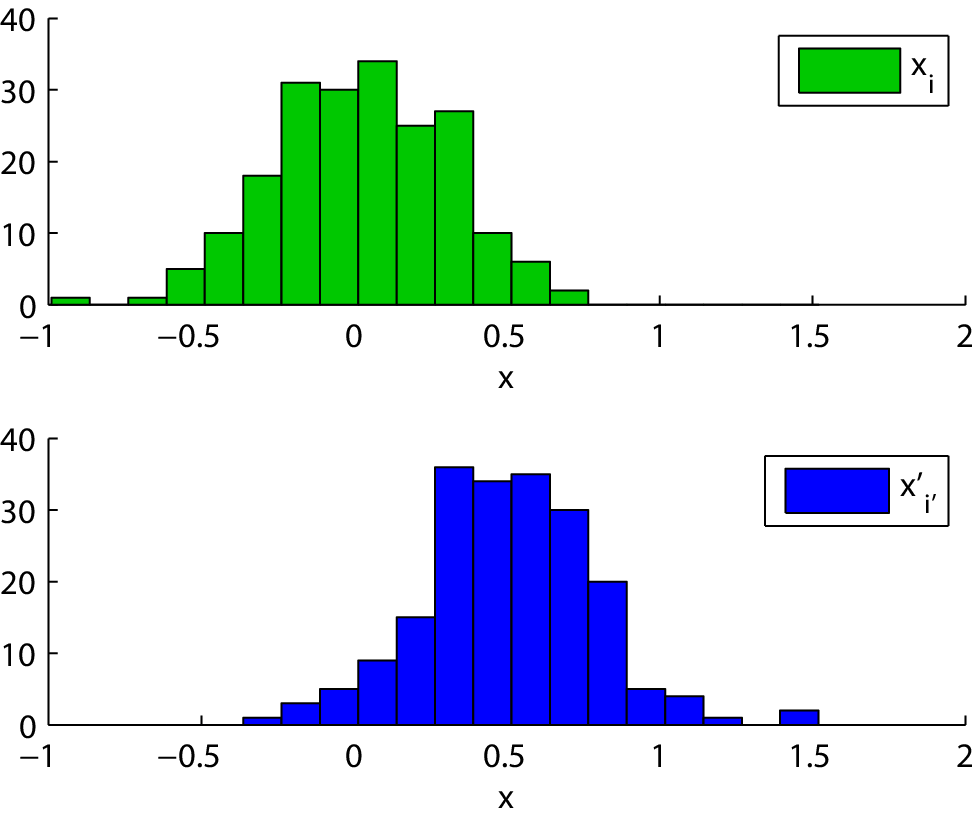}\\
    (a) Samples\\[4mm]
    \includegraphics[width=1\textwidth,clip]{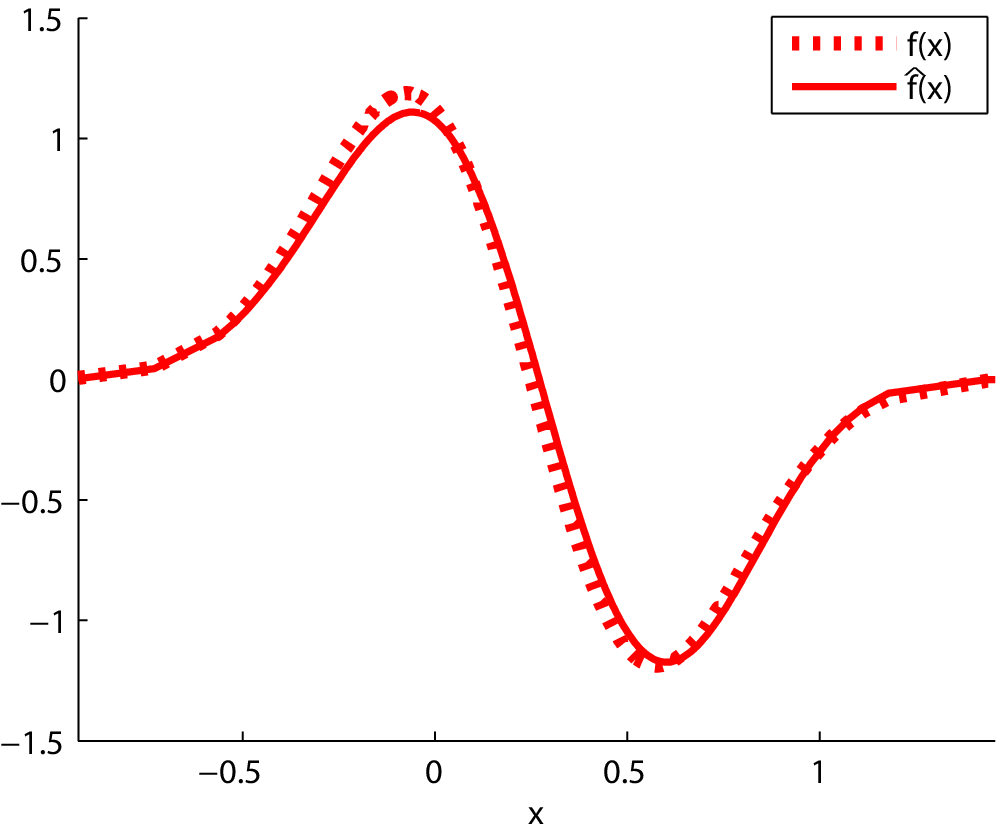}\\
    (b) LSDD\\[4mm]
    \includegraphics[width=1\textwidth,clip]{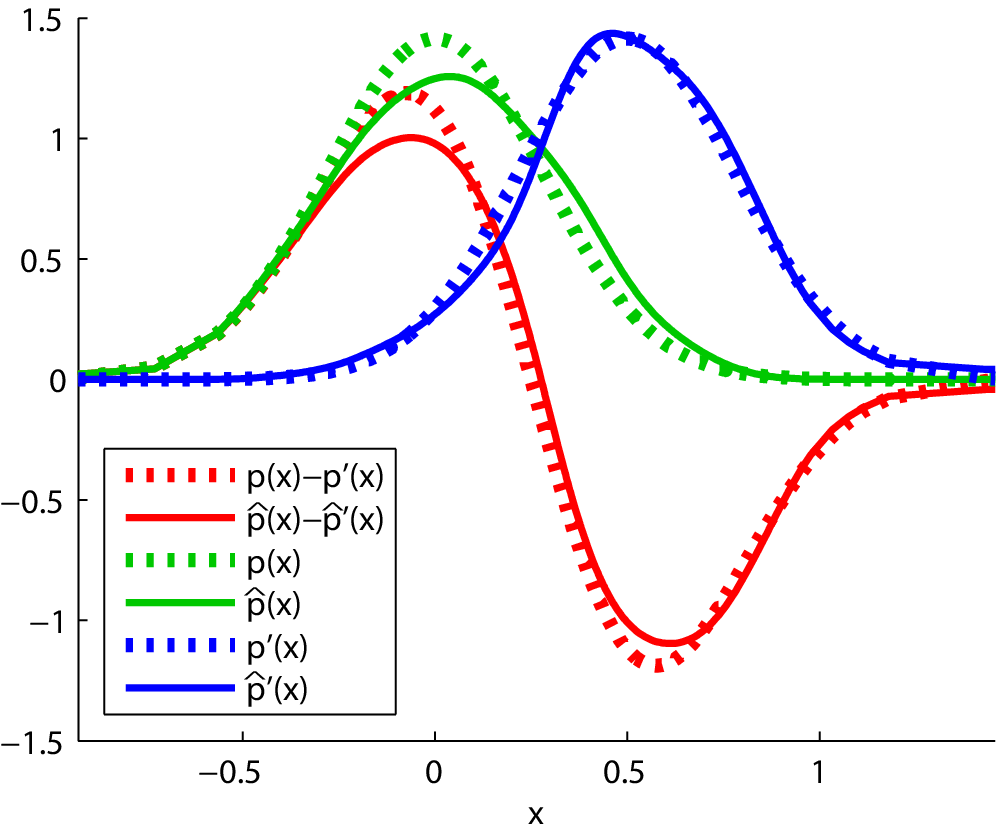}\\
    (c) KDE 
    \caption{
      Estimation of density difference when $\mu=0.5$ (i.e., $f(x)=\density(x)-\density'(x)\neq0$).
    }
    \label{fig:illustration-DD2-n200}
  \end{minipage}
\end{figure*}

% \begin{figure*}[t]
%   \centering
%   \subfigure[Samples]{
%     \includegraphics[width=0.31\textwidth,clip]{FIG/illustration-data-dataset1-d1-n200.eps}
%     }
%   \subfigure[LSDD]{
%     \includegraphics[width=0.31\textwidth,clip]{FIG/illustration-LSDD-dataset1-d1-n200.eps}
%   }
%   \subfigure[KDE]{
%     \includegraphics[width=0.31\textwidth,clip]{FIG/illustration-KDE-dataset1-d1-n200.eps}
%     }
%     \caption{
%     Estimation of density difference when $\mu=0$ (i.e., $f(x)=\density(x)-\density'(x)=0$).
%   }
%   \label{fig:illustration-DD1-n200}
% % \end{figure*}
% % \begin{figure*}[t]
% %   \centering
%   \vspace*{10mm}
%   \subfigure[Samples]{
%     \includegraphics[width=0.31\textwidth,clip]{FIG/illustration-data-dataset4-d1-n200.eps}
% }
%   \subfigure[LSDD]{
%     \includegraphics[width=0.31\textwidth,clip]{FIG/illustration-LSDD-dataset4-d1-n200.eps}
%     }
%   \subfigure[KDE]{
%     \includegraphics[width=0.31\textwidth,clip]{FIG/illustration-KDE-dataset4-d1-n200.eps}
%     }
%     \caption{
%       Estimation of density difference when $\mu=0.5$ (i.e., $f(x)=\density(x)-\density'(x)\neq0$).
%     }
%     \label{fig:illustration-DD2-n200}
% \end{figure*}

\subsubsection{LSDD vs.~KDE}\label{subsec:experiments-LSDDvsKDE}
We experimentally compare the behavior of LSDD and the KDE-based method.
Let
\begin{align*}
  \density(\boldx)&=N(\boldx;(\mu,0,\ldots,0)^\top,(4\pi)^{-1}\boldI_\inputdim),\\
  \density'(\boldx)&=N(\boldx;(0,0,\ldots,0)^\top,(4\pi)^{-1}\boldI_\inputdim),
\end{align*}
where $N(\boldx;\boldmu,\boldSigma)$ denotes 
the multi-dimensional normal density with 
mean vector $\boldmu$ and variance-covariance matrix $\boldSigma$ with respect to $\boldx$,
and $\boldI_\inputdim$ denotes the $\inputdim$-dimensional identity matrix.

\begin{figure*}[t]
  \centering
  \begin{tabular}{@{}cc@{}}
     \includegraphics[width=0.40\textwidth,clip]{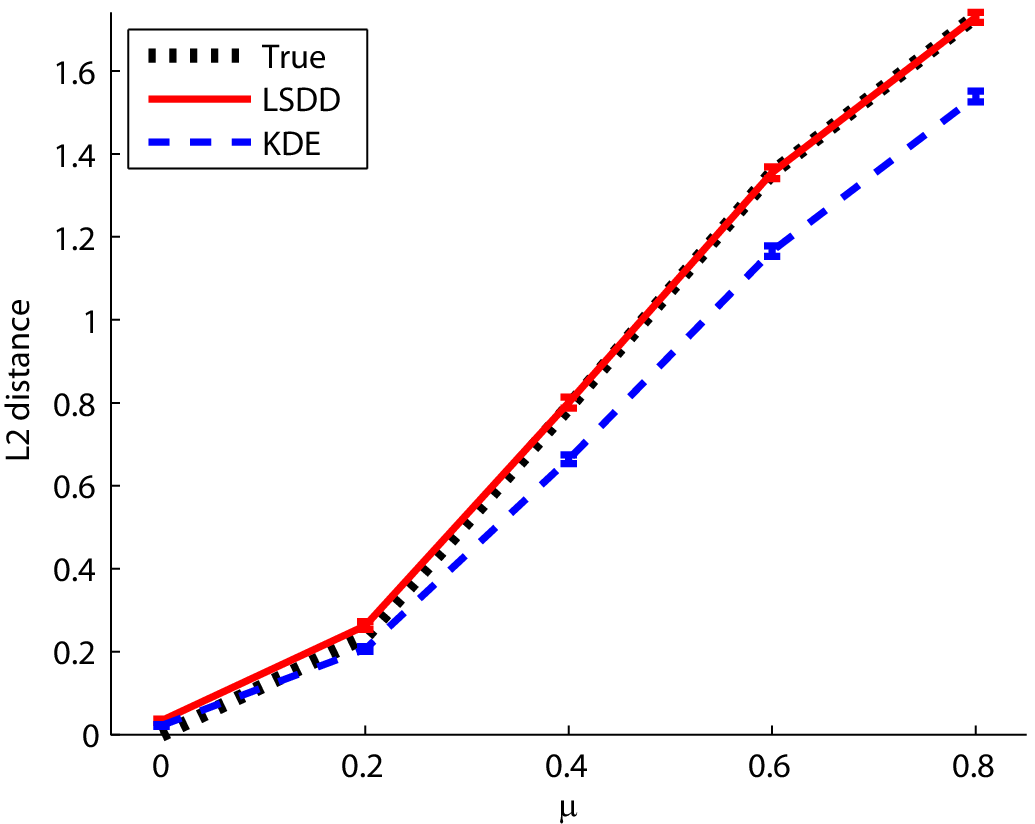}&
     \includegraphics[width=0.40\textwidth,clip]{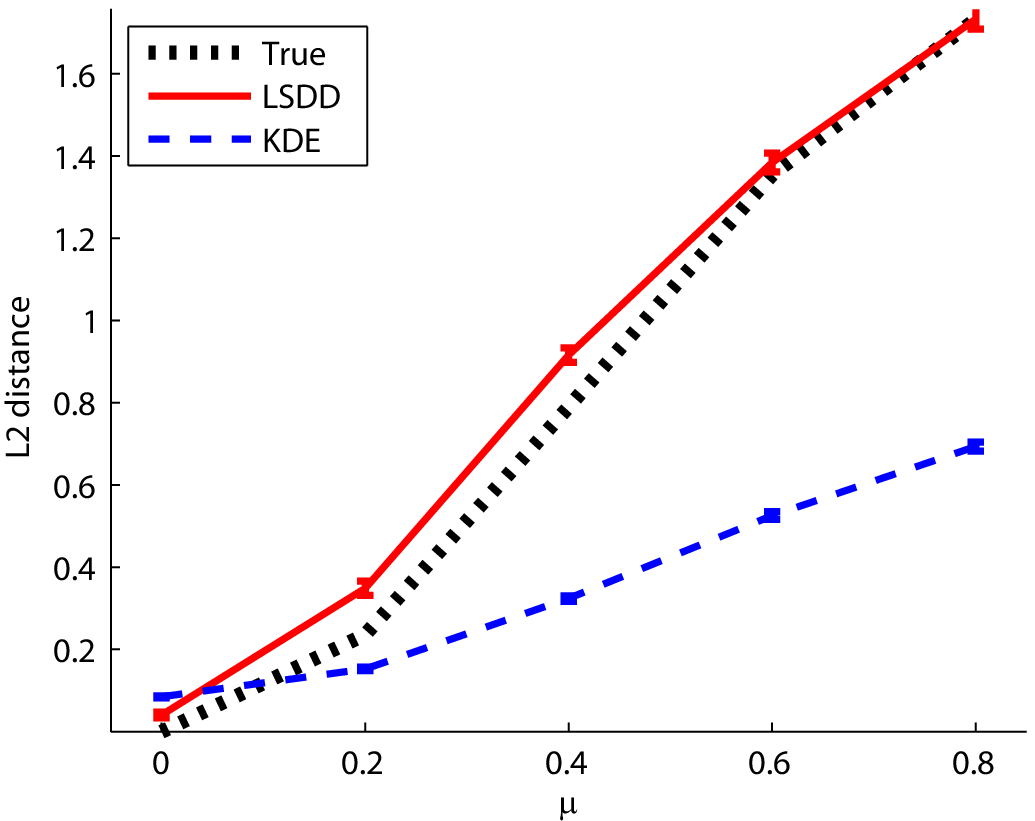}\\
    (a) $d=1$&
%    $d=3$&
    (b) $d=5$
  \end{tabular}
%   \subfigure[$d=1$]{
%     \includegraphics[width=0.22\textwidth,clip]{FIG/illustration-L2dist-d1-n200.eps}
%   \label{fig:illustration-L2dist-d1-n200}
%   }
%   \subfigure[$d=5$]{
%     \includegraphics[width=0.22\textwidth,clip]{FIG/illustration-L2dist-d5-n200.eps}
%   \label{fig:illustration-L2dist-d5-n200}
%   }
  \caption{
    $L^2$-distance estimation by LSDD and KDE.
    Means and standard errors over $100$ runs are plotted.
  }
  \label{fig:illustration-L2dist-n200}
\end{figure*}

We first illustrate how LSDD and KDE behave
under $\inputdim=1$ and $\nsample=\nsample'=200$.
Figure~\ref{fig:illustration-DD1-n200}
depicts the data samples, 
densities and density difference estimated by KDE,
and
density difference estimated by LSDD
for $\mu=0$ (i.e., $f(x)=\density(x)-\density'(x)=0$).
This shows that LSDD gives a more accurate estimate of the density difference $f(x)$
than KDE.
Figure~\ref{fig:illustration-DD2-n200}
depicts the results
for $\mu=0.5$ (i.e., $f(x)\neq 0$),
showing again that LSDD performs well.

Next, we compare the $L^2$-distance approximator based on LSDD 
and that based on KDE.
For $\mu=0,0.2,0.4,0.6,0.8$ and $\inputdim=1,5$, 
we draw $\nsample=\nsample'=200$ samples from
the above $\density(\boldx)$ and $\density'(\boldx)$.
Figure~\ref{fig:illustration-L2dist-n200} depicts
the mean and standard error of estimated $L^2$-distances over $100$ runs
as functions of mean $\mu$.
When $\inputdim=1$,
the LSDD-based $L^2$-distance estimator
gives accurate estimates of the true $L^2$-distance,
whereas the KDE-based $L^2$-distance estimator
slightly underestimates the true $L^2$-distance.
This is caused by the fact that KDE tends to provide smoother density estimates
(see Figure~\ref{fig:illustration-DD2-n200}(c) again).
% because the bias-variance trade-off is optimally controlled by cross-validation.
Such smoother density estimates are accurate as density estimates,
but the difference of smoother density estimates yields 
a smaller $L^2$-distance estimate \cite{JMA:Anderson+etal:1994}.
This tendency is more significant when $\inputdim=5$;
the KDE-based $L^2$-distance estimator severely underestimates the true $L^2$-distance,
which is a typical drawback of the two-step procedure.
On the other hand, the LSDD-based $L^2$-distance estimator
still gives reasonably accurate estimates
of the true $L^2$-distance even when $\inputdim=5$.
% However, we note that LSDD also slightly underestimates the true $L^2$-distance
% when $\inputdim=5$
% because the bias-variance trade-off is optimally controlled by cross-validation
% (see Section~\ref{sec:LSDD-CV}).

\subsubsection{$L^2$-Distance vs.~KL-Divergence}
The \emph{Kullback-Leibler} (KL) divergence
\cite{Annals-Math-Stat:Kullback+Leibler:1951}
is a popular divergence measure for comparing probability distributions.
The KL-divergence from $\density(\boldx)$ to $\density'(\boldx)$ is defined as
\begin{align*}
\mathrm{KL}(\density\|\density')
:=  \int \density(\boldx)\log\frac{\density(\boldx)}{\density'(\boldx)}\mathrm{d}\boldx.
\end{align*}

\begin{figure}[p]
\centering
    \includegraphics[width=0.40\textwidth,clip]{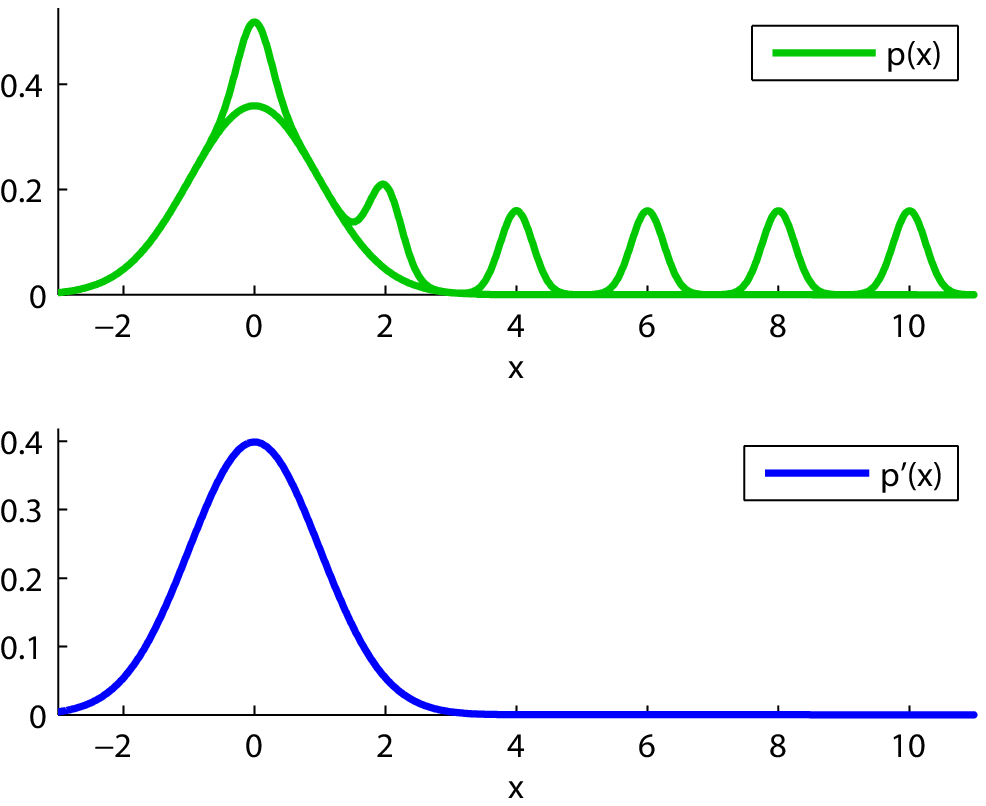}
\vspace*{-3mm}
  \caption{
    Comparing two densities in the presence of outliers.
    $\density(x)$ includes outliers at $\mu=0,2,4,\ldots,10$.
  }
  \label{fig:exp-illustrative-densities}
\vspace*{5mm}
  \begin{minipage}[t]{0.48\textwidth}
  \centering
    \includegraphics[width=0.8\textwidth,clip]{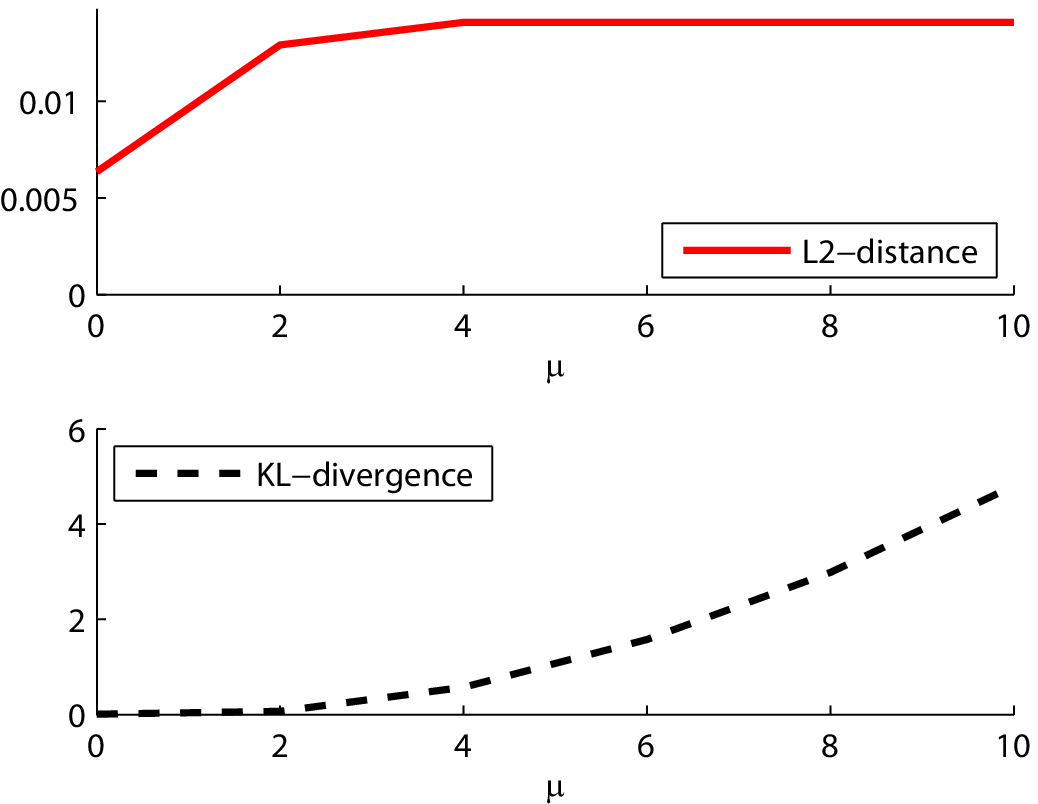}
\vspace*{-3mm}
  \caption{
    The true $L^2$-distance and true KL-divergence
    as functions of outlier mean $\mu$.
  }
  \label{fig:exp-illustrative-KL-L2}
\end{minipage}
~~~
  \begin{minipage}[t]{0.48\textwidth}
  \centering
    \includegraphics[width=0.8\textwidth,clip]{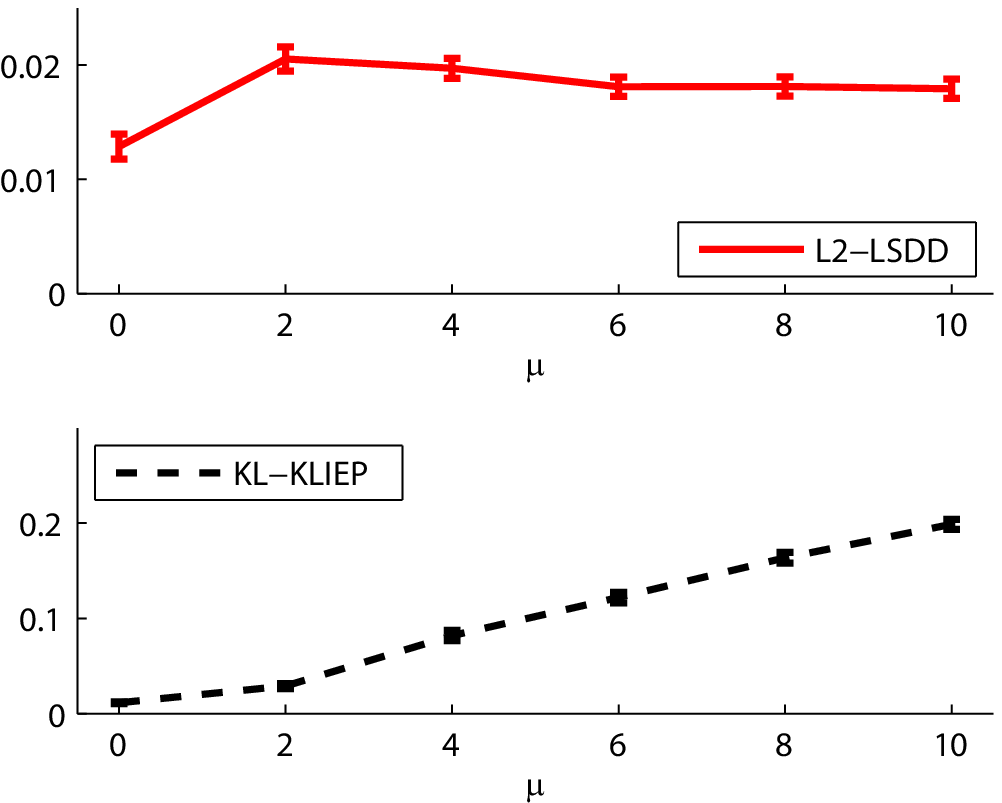}
\vspace*{-3mm}
  \caption{
    Means and standard errors of $L^2$-distance estimation by LSDD
    and KL-divergence estimation by KLIEP over $100$ runs.
%     The scale of KLIEP values is much different from the true values
%     (see Figure~\ref{fig:exp-illustrative-KL-L2})
%     because empirical normalization was unreliable.
%  (see Eq.\eqref{KLIEP-constraint}).
  }
  \label{fig:exp-illustrative-LSDD-KLIEP}
\end{minipage}
% \end{figure}
% \begin{figure}[t]
%   \centering
\vspace*{5mm}
  \begin{minipage}[t]{0.48\textwidth}
  \centering
    \includegraphics[width=0.8\textwidth,clip]{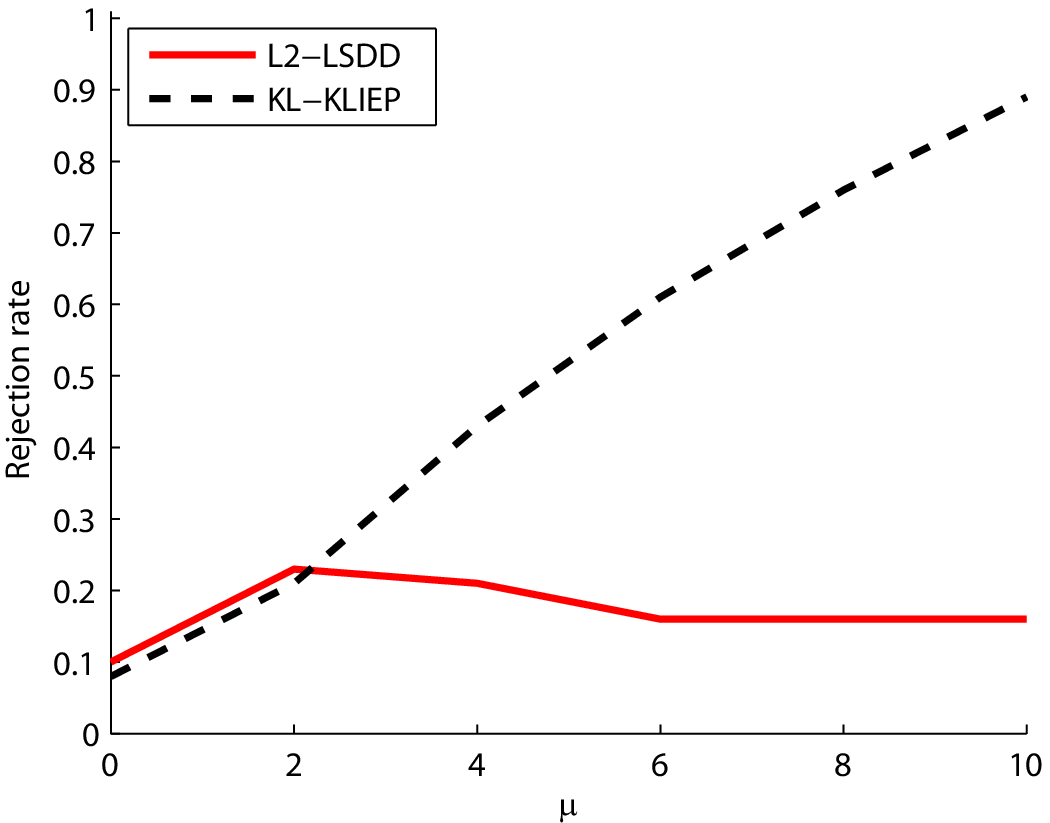}
  \caption{
    Two-sample test for outlier rate $\eta=0.1$
    as functions of outlier mean $\mu$.
  }
  \label{fig:exp-illustrative-2sampletest}
\end{minipage}
~~~
  \begin{minipage}[t]{0.48\textwidth}
  \centering
    \includegraphics[width=0.8\textwidth,clip]{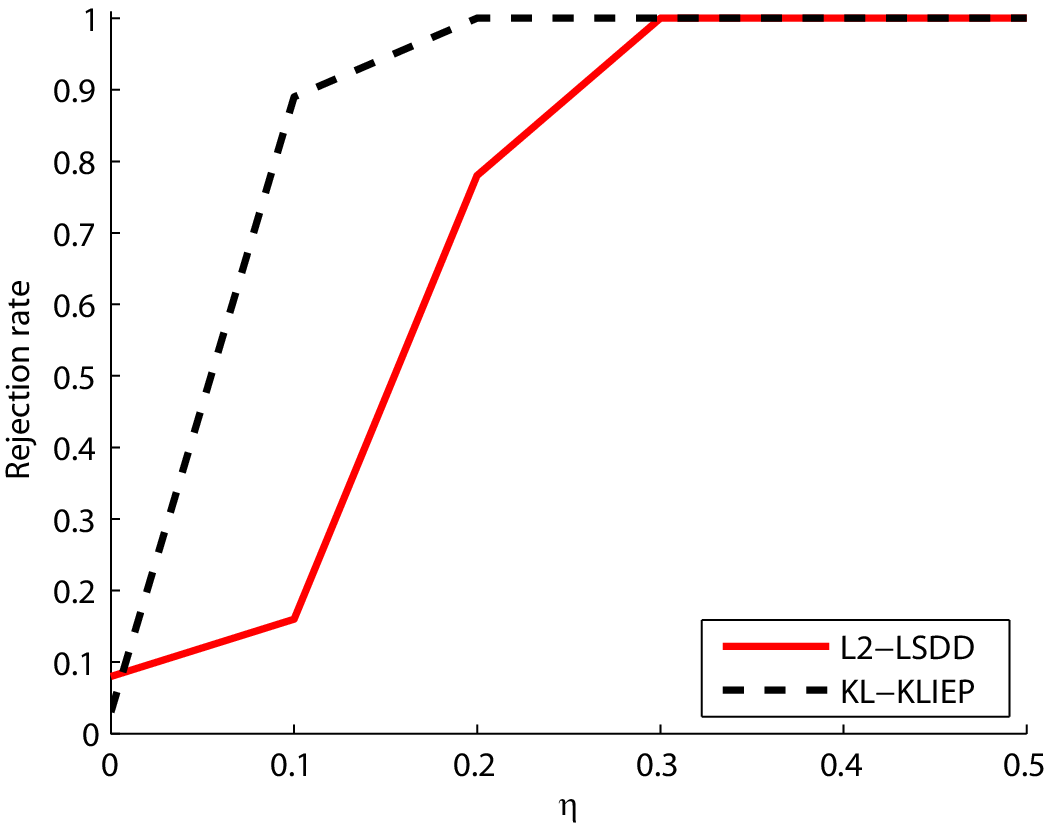}
  \caption{
    Two-sample test for outlier mean $\mu=10$
    as functions of outlier rate $\eta$.
  }
  \label{fig:exp-illustrative-2sampletest-mu}
\end{minipage}
\end{figure}

First, we illustrate the difference between the $L^2$-distance and the KL-divergence.
For $\inputdim=1$, let
\begin{align*}
  \density(x)&=(1-\eta)N(x;0,1^2)+\eta N(x;\mu,1/4^2),\\
  \density'(x)&=N(x;0,1^2).
\end{align*}
% where $N(x,\mu,\sigma^2)$ denotes the probability density function
% of the normal distribution with mean $\mu$ and variance $\sigma^2$ with respect to $x$.
Implications of the above densities are that 
samples drawn from $N(x;0,1^2)$ are inliers,
whereas samples drawn from $N(x;\mu,1/4^2)$ are outliers.
We set the outlier rate at $\eta=0.1$
and the outlier mean at $\mu=0,2,4,\ldots,10$
(see Figure~\ref{fig:exp-illustrative-densities}).

Figure~\ref{fig:exp-illustrative-KL-L2} depicts
the $L^2$-distance and the KL-divergence for outlier mean $\mu=0,2,4,\ldots,10$.
This shows that both the $L^2$-distance and the KL-divergence increase as $\mu$ increases.
However, the $L^2$-distance is bounded from above,
whereas the KL-divergence diverges to infinity as $\mu$ tends to infinity.
This result implies that the $L^2$-distance is less sensitive to outliers
than the KL-divergence,
which well agrees with the observation given in
\emcite{Biometrika:Basu+etal:1998}.

Next, we draw $\nsample=\nsample'=100$ samples from $\density(x)$ and $\density'(x)$,
and estimate the $L^2$-distance by LSDD and the KL-divergence
by the \emph{Kullback-Leibler importance estimation procedure}\footnote{
Estimation of the KL-divergence from data has been extensively studied recently
\cite{IEEE-IT:Wang+etal:2005,AISM:Sugiyama+etal:2008,ISIT:Perez-Cruz:2008,JSPI:Silva+Narayanan:2010,IEEE-IT:Nguyen+etal:2010}.
Among them, KLIEP was shown to possess a superior convergence property
and demonstrated to work well in practice.
KLIEP is based on direct estimation of density ratio
$\density(\boldx)/\density'(\boldx)$
without density estimation of $\density(\boldx)$ and $\density'(\boldx)$.
} 
(KLIEP)
\cite{AISM:Sugiyama+etal:2008,IEEE-IT:Nguyen+etal:2010}.
%; see also Appendix~\ref{sec:KLIEP},
Figure~\ref{fig:exp-illustrative-LSDD-KLIEP} depicts
the estimated $L^2$-distance and KL-divergence for outlier mean $\mu=0,2,4,\ldots,10$
over $100$ runs.
This shows that both LSDD and KLIEP reasonably capture
the profiles of the true $L^2$-distance and the KL-divergence,
although the scale of KLIEP values is much different from the true values
(see Figure~\ref{fig:exp-illustrative-KL-L2})
because the estimated normalization factor was unreliable.
%(see Eq.\eqref{KLIEP-constraint}).
% % nevertheless, the profile of KLIEP captures the true KL-divergence well.

Finally, based on the \emph{permutation test} procedure \cite{book:Efron+Tibshirani:1993},
we conduct hypothesis testing of the null hypothesis
that densities $\density$ and $\density'$ are the same.
More specifically, we first compute a distance estimator
for the original datasets $\calX$ and $\calX'$
and obtain $\widehat{D}(\calX,\calX')$.
Next, we randomly permute the $|\calX\cup\calX'|$ samples,
and assign the first $|\calX|$ samples to a set $\calXt$
and the remaining $|\calX'|$ samples to another set $\calXt'$.
Then we compute the distance estimator 
again using the randomly permuted datasets $\calXt$ and $\calXt'$
and obtain $\widetilde{D}(\calXt,\calXt')$.
Since $\calXt$ and $\calXt'$ can be regarded as being drawn from the same distribution,
$\widetilde{D}(\calXt,\calXt')$ would take a value close to zero.
This random permutation procedure is repeated many times,
and the distribution of $\widetilde{D}(\calXt,\calXt')$
under the null hypothesis (i.e., the two distributions are the same) is constructed.
Finally, the p-value is approximated by evaluating the relative ranking of
$\widehat{D}(\calX,\calX')$ in the histogram of $\widetilde{D}(\calXt,\calXt')$.
We set the significance level at $5\%$.

Figure~\ref{fig:exp-illustrative-2sampletest}
depicts the rejection rate of the null hypothesis for
outlier rate $\eta=0.1$ and outlier mean $\mu=0,2,4,\ldots,10$,
based on the $L^2$-distance estimated by LSDD and
the KL-divergence estimated by KLIEP.
This shows that the KLIEP-based test rejects the null hypothesis more frequently
for large $\mu$, whereas the rejection rate of the LSDD-based test
is kept almost constant even when $\mu$ is changed.
This result implies that the two-sample test by LSDD is more robust
against outliers (i.e., two distributions tend to be regarded as the same
even in the presence of outliers) than the KLIEP-based test.

Figure~\ref{fig:exp-illustrative-2sampletest-mu}
depicts the rejection rate of the null hypothesis for outlier mean $\mu=10$
for outlier rate $\eta=0,0.05,0.1,\ldots,0.35$.
When $\eta=0$ (i.e., no outliers), 
both the LSDD-based test and the KLIEP-based test
accept the null hypothesis with the designated significance level approximately.
When $\eta=0.1$,
the LSDD-based test still keeps a low rejection rate,
whereas the KLIEP-based test tends to reject the null hypothesis.
When $\eta\ge0.3$, the LSDD-based test and the KLIEP-based test
tend to reject the null hypothesis in a similar way.

% Next, we conduct a similar experiment for the USPS dataset
% that consists of hand-written digit images of $16\times16$ pixels
% (i.e., $256$ dimensions).
% The first dataset consists of $300$ images of digit ``0'',
% whereas the second dataset consists of $300(1-\eta)$ images of digit ``0''
% and $300\eta$ images of digit ``1'' for $\eta=0,0.02,0.04,\ldots,0.12$.
% The images of digit ``1'' are regarded as outliers here.
% Figure~\ref{fig:exp-USPS-2sampletest-dataset=1}
% depicts the rejection rate of the null hypothesis
% for the USPS dataset, showing again that the LSDD-based test
% tends to be more robust against outliers than the KLIEP-based test.

% \begin{figure}[t]
%   \centering
%     \includegraphics[width=0.40\textwidth,clip]{FIG/USPS-2sampletest-dataset=1.eps}
%   \caption{
%     Two-sample test for the USPS dataset
%     as functions of outlier rate $\eta$.
%   }
%   \label{fig:exp-USPS-2sampletest-dataset=1}
% \end{figure}

\begin{figure}[t]
  \centering
  \includegraphics[width=0.45\textwidth,clip]{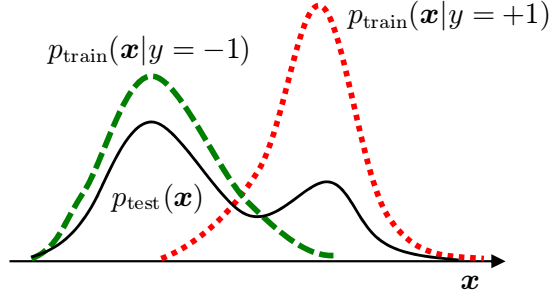}
  \caption{Schematic illustration of semi-supervised class-balance estimation.}
  \label{fig:class-balance-notation}
\end{figure}

\begin{figure}[p]
  \centering
 \subfigure[Australian dataset]{
      \includegraphics[width=0.4\textwidth,clip]{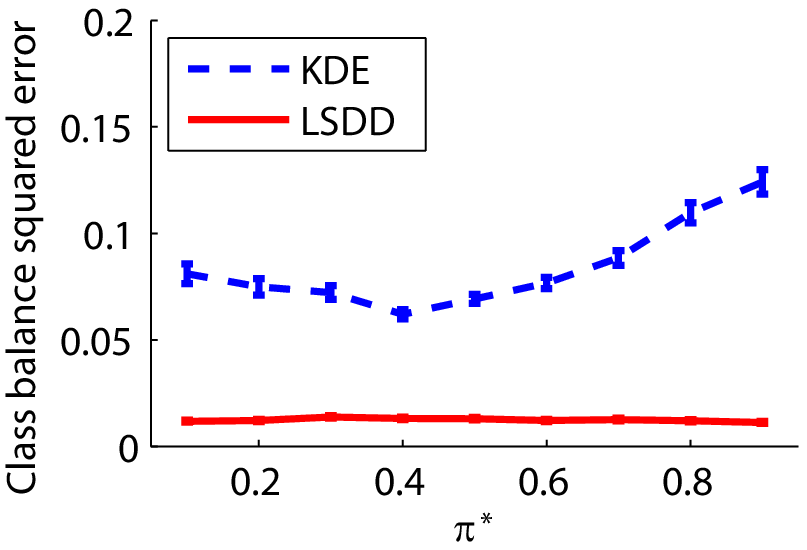}
   \includegraphics[width=0.4\textwidth,clip]{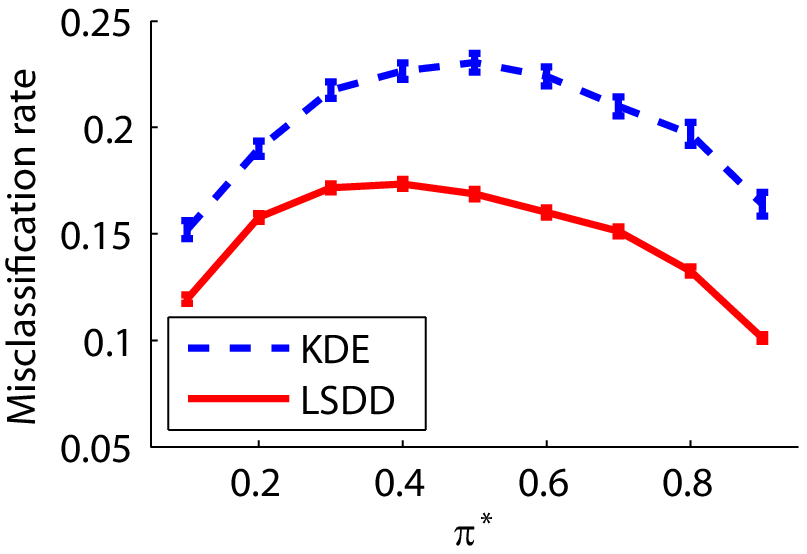}
  \label{fig:class-balance-australian}
  }
  \subfigure[Diabetes dataset]{
      \includegraphics[width=0.4\textwidth,clip]{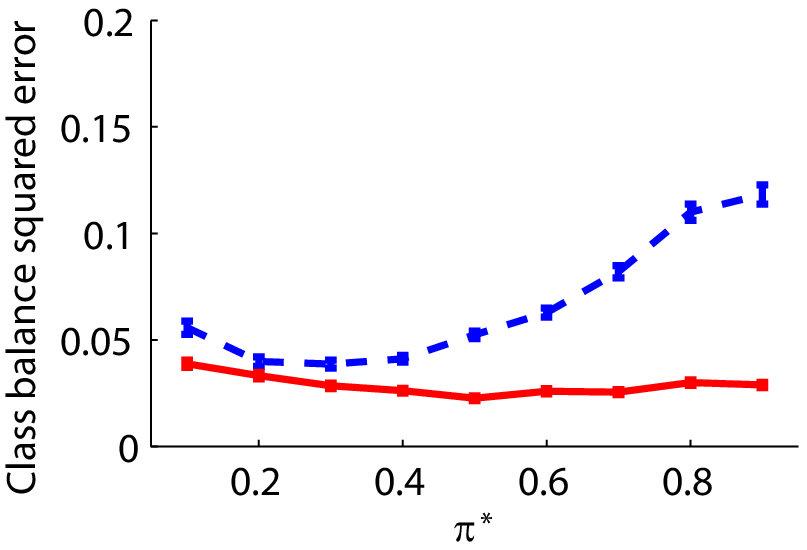}
   \includegraphics[width=0.4\textwidth,clip]{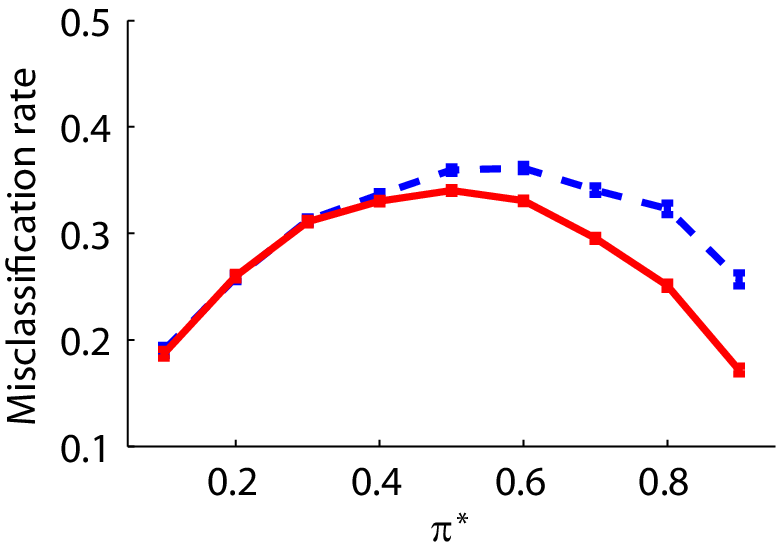}
  \label{fig:class-balance-diabetes}
  }\\
  \subfigure[German dataset]{
      \includegraphics[width=0.4\textwidth,clip]{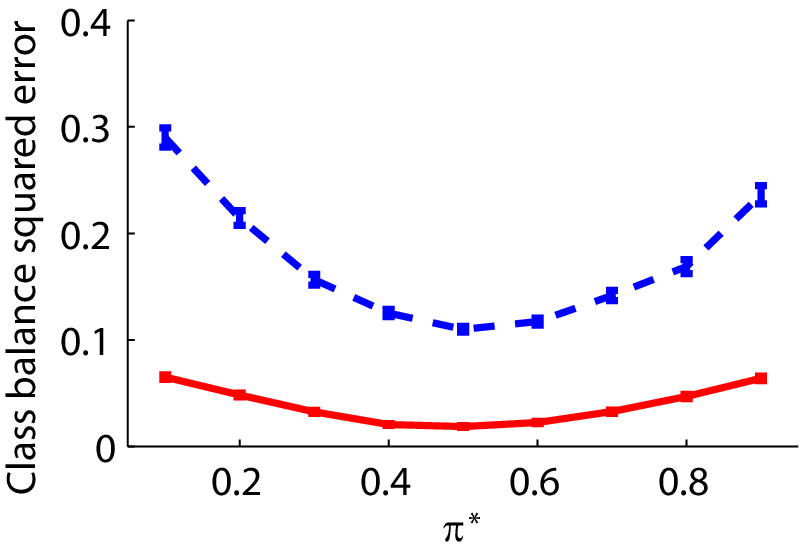}
   \includegraphics[width=0.4\textwidth,clip]{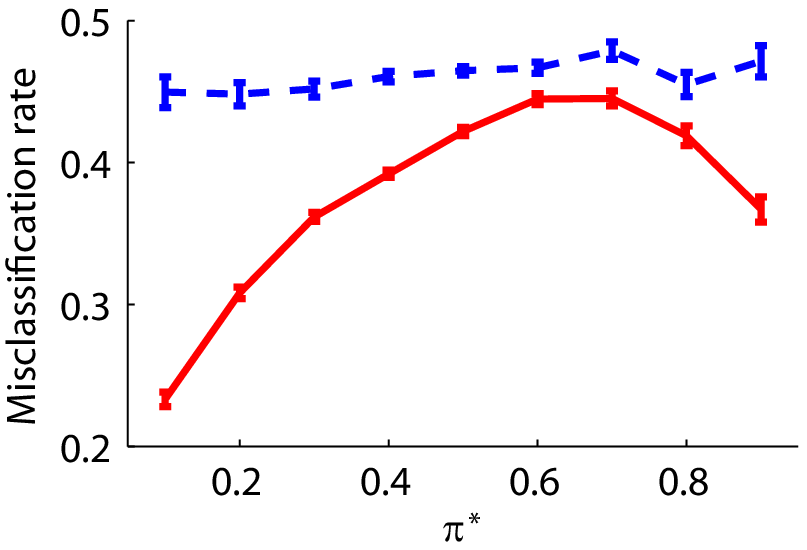}
  \label{fig:class-balance-german}
  }
%   \subfigure[Ionosphere dataset]{
%     \begin{tabular}{@{}c@{}}
%       \includegraphics[width=0.4\textwidth,clip]{FIG/class-balance-ionosphere.eps}\\
%    \includegraphics[width=0.4\textwidth,clip]{FIG/class-balance-ionosphere-Misclass.eps}
%     \end{tabular}
%   \label{fig:class-balance-ionosphere}
%   }
  \subfigure[Statlogheart dataset]{
      \includegraphics[width=0.4\textwidth,clip]{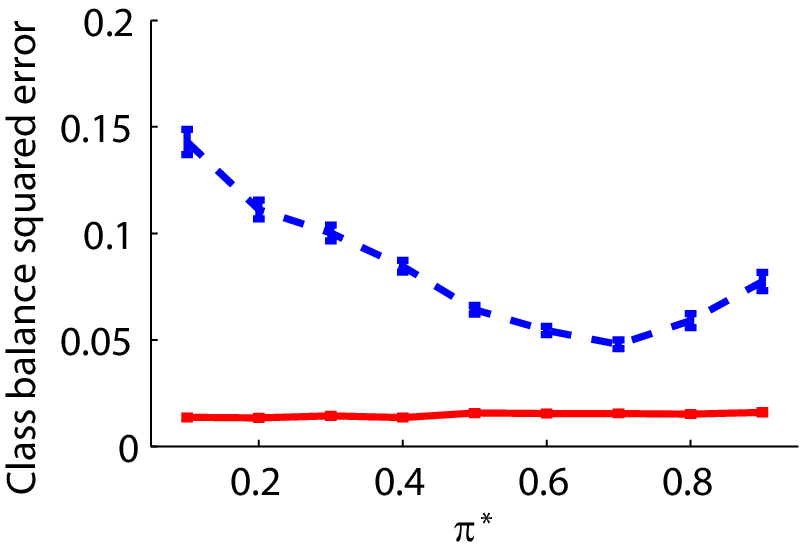}
   \includegraphics[width=0.4\textwidth,clip]{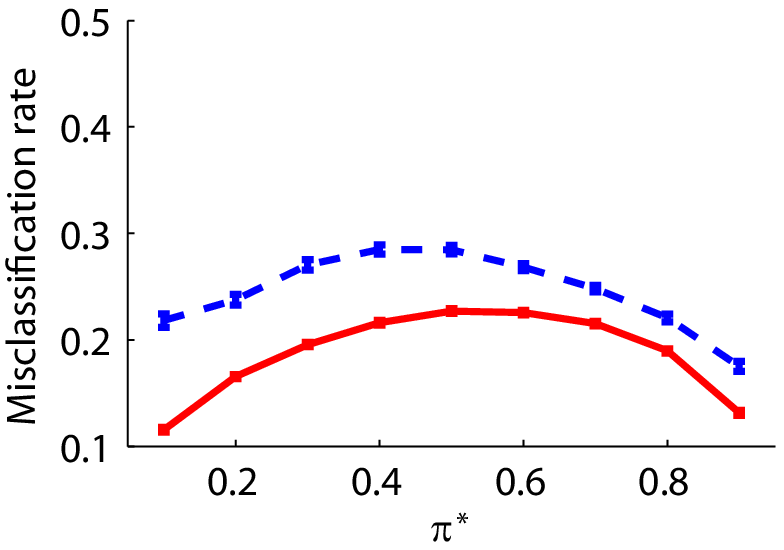}
  \label{fig:class-balance-statlogheart}
  }
  \caption{Results of semi-supervised class-balance estimation.
    Left: Squared error of class balance estimation.
    Right: Misclassification error
  by a weighted regularized least-squares classifier.}
  \label{fig:class-balance}
\end{figure}

\subsection{Applications}
\label{subsec:applications}
Next, we apply LSDD to
semi-supervised class-balance estimation under class prior change
and change-point detection in time series.

\subsubsection{Semi-Supervised Class-Balance Estimation}
In real-world pattern recognition tasks, changes in class balance are
often observed. Then significant estimation bias can be caused
since the class balance in the training dataset does not reflect that
of the test dataset. 

Here, we consider a pattern recognition task of
classifying pattern $\boldx\in\mathbbR^d$ to class $y\in\{+1,-1\}$.
Our goal is to learn the class balance of a test dataset
in a semi-supervised learning setup where
unlabeled test samples are provided in addition to labeled training samples
\cite{book:Chapelle+etal:2006}.
The class balance in the test set can be estimated
by matching a mixture of class-wise training input densities,
\[
\pi p_\mathrm{train}(\boldx|y=+1)+(1-\pi)p_\mathrm{train}(\boldx|y=-1),
\]
with the test input density $p_\mathrm{test}(\boldx)$
\cite{nc:Saerens+Latinne+Decaestecker:2002},
where $\pi\in[0,1]$ is a mixing coefficient to learn.
See Figure~\ref{fig:class-balance-notation} for schematic illustration.
Here, we use the $L^2$-distance estimated by LSDD and the difference of KDEs
for this distribution matching.

We use four UCI benchmark datasets\footnote{\url{http://archive.ics.uci.edu/ml/}},
where we randomly choose $20$ labeled training samples from each class
and $50$ unlabeled test samples following true class-prior $\pi^*=0.1,0.2,\ldots,0.9$.
Figure~\ref{fig:class-balance}
plots the mean and standard error of the squared difference between
true and estimated class balances $\pi$
and the misclassification error by a weighted regularized least-squares classifier
\cite{NATO-ASI:Rifkin+Yeo+Poggio:2003} over $1000$ runs.
The results show that LSDD tends to provide better class-balance estimates,
which are translated into lower classification errors.

\begin{figure}[t]
  \centering
   \includegraphics[width=0.65\textwidth,clip]{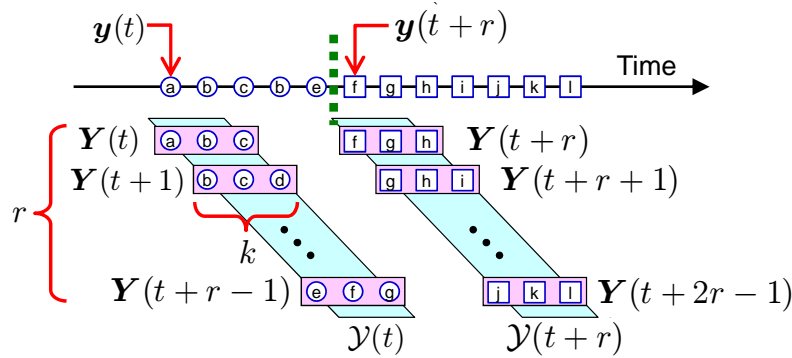}
   \caption{Schematic illustration of unsupervised change detection.}
  \label{fig:change-detection-notation}
\end{figure}
\begin{figure}[p]
  \centering
  \subfigure[Speech data]{
   \includegraphics[width=0.7\textwidth,clip]{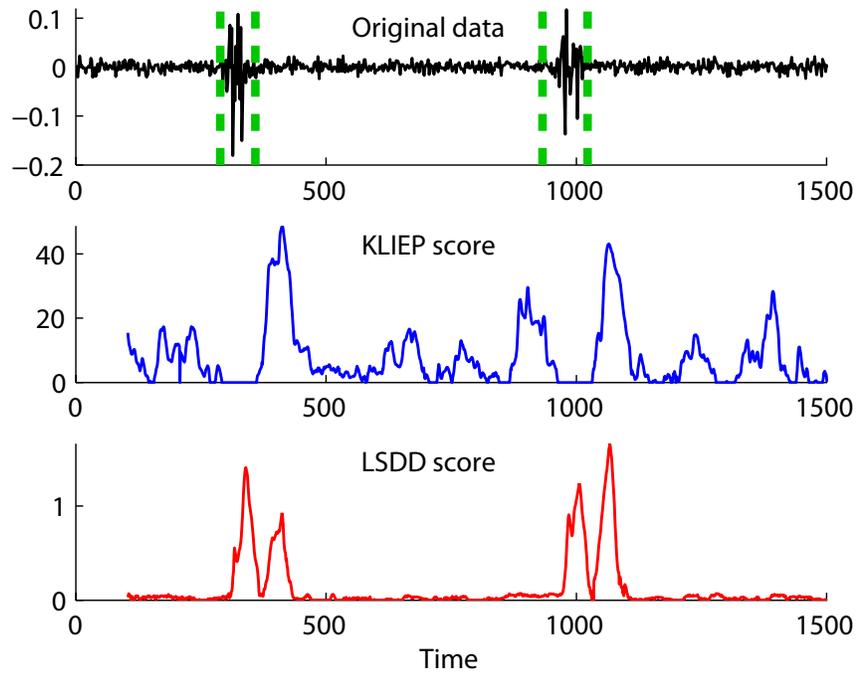}
  \label{fig:change-detection-speech}
  }
  \subfigure[Accelerometer data]{
   \includegraphics[width=0.7\textwidth,clip]{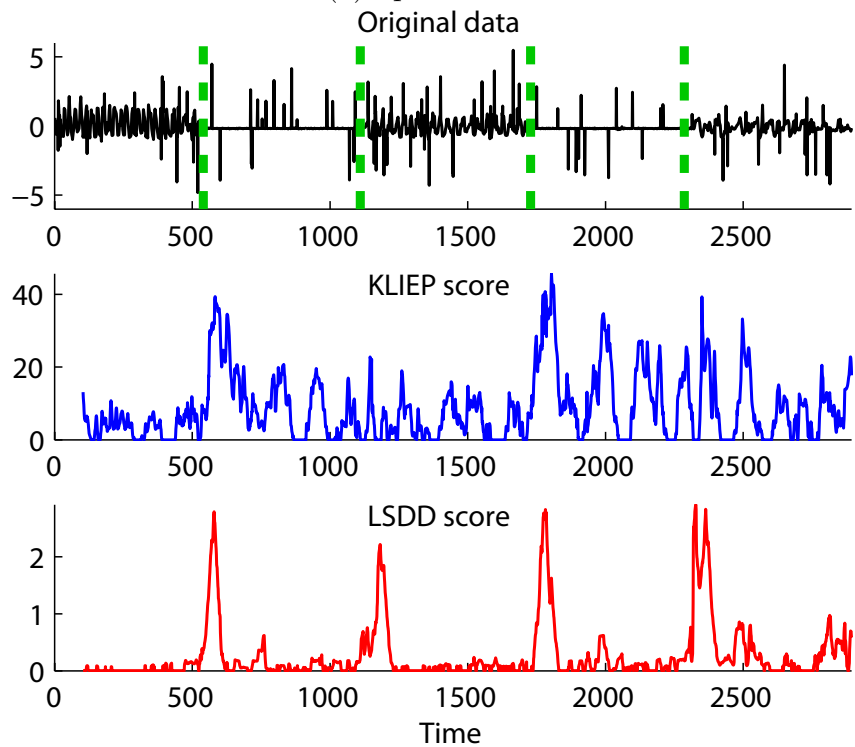}
  \label{fig:change-detection-accelerometer}
  }
  \caption{Results of unsupervised change detection.
  Top: Original time-series.
  Middle: Change scores obtained by KLIEP.
  Bottom: Change scores obtained by LSDD.
  }
  \label{fig:change-detection}
\end{figure}

\subsubsection{Unsupervised Change Detection}
The objective of change detection is to discover abrupt property
changes behind time-series data. 

Let $\boldy(t)\in \mathbbR^m$
be an $m$-dimensional time-series sample at time $t$,
and let
\[
\boldY(t) := [ \boldy(t)^\top, \boldy(t+1)^\top, \ldots,\boldy(t+k-1)^\top]^\top
\in \mathbb{R}^{km}
\]
be a subsequence of time series at time $t$ with length $k$.
We treat the subsequence $\boldY(t)$ as a sample,
instead of a single point $\boldy(t)$,
by which time-dependent information can be incorporated naturally
\cite{SAM:Kawahara+Sugiyama:2012}.
Let $\calY(t)$ be a set of $r$ retrospective subsequence samples
starting at time $t$:
\begin{align*}
&\calY(t) := \{{\boldY(t)},\boldY(t+1),\hdots,\boldY(t+r-1)\}.
\end{align*}
Our strategy is to compute a certain dissimilarity measure
between two consecutive segments $\calY(t)$ and $\calY(t+r)$,
and use it as the plausibility of change points
(see Figure~\ref{fig:change-detection-notation}).
As a dissimilarity measure,
we use the $L^2$-distance estimated by LSDD and the KL-divergence
estimated by the \emph{KL importance estimation procedure} (KLIEP)
\cite{AISM:Sugiyama+etal:2008,IEEE-IT:Nguyen+etal:2010}.
We set $k=5$ and $r=50$.

% A toy time-series data is plotted
% in the top graph in Figure~\ref{fig:change-detection-toy},
% where a change point exists at time $600$ (shown by green vertical lines)
% and outliers exist between time $300$ and $600$.
% The bottom two graphs in Figure~\ref{fig:change-detection-toy} plot
% change scores obtained by KLIEP and LSDD,
% showing that the LSDD score is much smoother and more robust
% against outliers than the KLIEP score.
% Thus, LSDD may produce less false alarms.

First,
we use the \emph{IPSJ SIG-SLP Corpora and Environments for Noisy Speech Recognition} (CENSREC)
dataset\footnote{\url{http://research.nii.ac.jp/src/en/CENSREC-1-C.html}}
provided by the \emph{National Institute of Informatics, Japan},
which records human voice in a noisy environment such as a restaurant.
The top graph in Figure~\ref{fig:change-detection-speech}
displays the original time-series,
where true change points were manually annotated.
The bottom two graphs in Figure~\ref{fig:change-detection-speech} plot
change scores obtained by KLIEP and LSDD,
showing that the LSDD-based change score indicates the existence of change points
more clearly than the KLIEP-based change score.

Next,
we use a dataset taken from the \emph{Human Activity Sensing Consortium (HASC)
challenge 2011}\footnote{\url{http://hasc.jp/hc2011/}},
which provides human activity information collected by
portable three-axis accelerometers.
%  The task of change-point detection
% is to segment the time-series data according to the behaviors. There
% are 6 behaviors (``stay'', ``walk'', ``jog'', ``skip'', ``stair up'',
% and ``stair down'') within a single time-series data.
%  The starting
% time of each behavior is arbitrarily decided by each user. 
Because the orientation of the accelerometers is not necessarily fixed, we take
the $\ell_2$-norm of the 3-dimensional data.
The top graph in Figure~\ref{fig:change-detection-accelerometer}
displays the original time-series
for a sequence of actions ``jog'', ``stay'', ``stair down'', ``stay'', and ``stair up''
(there exists $4$ change points at time $540$, $1110$, $1728$, and $2286$).
The bottom two graphs in Figure~\ref{fig:change-detection-accelerometer} depict
the change scores obtained by KLIEP and LSDD,
showing that the LSDD score is much more stable 
and interpretable than the KLIEP score.

% ===============================

% Figure~\ref{fig:exp-illustrative-2sampletest-eta}
% depicts the rejection rate of the null hypothesis for 
% outlier mean $\mu=0,2,4,\ldots,10$
% with outlier rate $\eta=0,0.1,0.2,\ldots,0.5$.
% When $\eta=0$ (i.e., no outliers), 
% both the LSDD test and the KLIEP test
% accept the null hypothesis with the designated significance level.
% When $\eta=0.1$,
% the LSDD test still accepts the null hypothesis rather frequently
% for all $\mu$,
% whereas the KLIEP test tends to reject it for large $\mu$.
% This demonstrates the robustness of the LSDD test against outliers.
% When $\eta\ge0.2$, the LSDD test and the KLIEP test
% tend to reject the null hypothesis in the same way.

% ===============================

% \begin{figure}[t]
%   \centering
%   \subfigure[LSDD]{
%     \includegraphics[width=0.45\textwidth,clip]{FIG/illustration-2sampletest-LSDD.eps}
% }
%   \subfigure[KLIEP]{
%     \includegraphics[width=0.45\textwidth,clip]{FIG/illustration-2sampletest-KLIEP.eps}
% }
%   \caption{
%     Two-sample test for illustrative data
%     with outlier rate $\eta=0,0.05,0.1,0.15,0.2$.
%     When $\eta=0.1$, the results are the same as 
%     those plotted in Figure~\ref{fig:exp-illustrative-2sampletest}.
%   }
%   \label{fig:exp-illustrative-2sampletest-eta}
% \end{figure}

%%%%%%%%%%%%%%%%%%%%%%%%%%%%%%%%%%%%%%%%%%%%%%%%%%%%%%%%%%%%
\section{Conclusions}
\label{sec:conclusion}

In this paper, we proposed a method for directly estimating
the difference between two probability density functions
without density estimation.
The proposed method, called the \emph{least-squares density-difference} (LSDD),
was derived within a framework of kernel least-squares estimation,
and its solution can be computed analytically in a computationally efficient
and stable manner.
Furthermore, LSDD is equipped with cross-validation,
and thus all tuning parameters such as the kernel width
and the regularization parameter can be systematically and objectively optimized.
We showed the asymptotic normality of LSDD in a parametric setup
and derived a finite-sample error bound
for LSDD in a non-parametric setup.
In both cases, LSDD achieves the optimal convergence rate.

We also proposed an $L^2$-distance estimator based on LSDD,
which nicely cancels a bias caused by regularization.
The LSDD-based $L^2$-distance estimator
was experimentally shown to be more accurate than the difference of kernel density estimators
and more robust against outliers than Kullback-Leibler divergence estimation.

Density-difference estimation is a novel research paradigm in machine learning,
and we have given a simple but useful method for this emerging topic.
Our future work will develop more powerful algorithms for density-difference estimation
and explores a variety of applications.

%%%%%%%%%%%%%%%%%%%%%%%%%%%%%%%%%%%%%%%%%%%%%%%%%%%%%%%%%%%%
\section*{Acknowledgments}
The authors would like to thank Wittawat Jitkrittum for his comments.
Masashi Sugiyama was supported by MEXT KAKENHI 23300069,
Takafumi Kanamori was supported by MEXT KAKENHI 24500340,
Taiji Suzuki was supported by MEXT KAKENHI 22700289
and the Aihara Project, the FIRST program from JSPS initiated by CSTP,
Marthinus Christoffel du Plessis was supported by MEXT Scholarship,
Song Liu was supported by the JST PRESTO program,
and
Ichiro Takeuchi was supported by MEXT KAKENHI 23700165.

%%%%%%%%%%%%%%%%%%%%%%%%%%%%%%%%%%%%%%%%%%%%%%%%%%%%%%%%%%%%
\bibliography{E:/work/bib/mybib,E:/work/bib/sugiyama}

\appendix
\section{Technical Details of Non-Parametric Convergence Analysis
  in Section~\ref{subsec:nonparametric-convergence}}
\label{proof-th:TheMainBound}

First, we define linear operators $P_n,P,P'_n,P',Q_n,Q$ as 
\begin{align*}
P_n f := \frac{1}{n}\sum_{i=1}^n f(\boldx_i),~~
&P f := \int_{\Real^d} f(\boldx) p(\boldx) \dd \boldx, \\
P'_n f := \frac{1}{n}\sum_{i=1}^n f(\boldx'_i),~~
&P' f := \int_{\Real^d} f(\boldx) p'(\boldx) \dd \boldx, \\
Q_n f := P_n f - P'_n f,~~
&Q f := P f - P' f.
\end{align*}
Let $\calHg$ be an RKHS endowed with the Gaussian kernel with width $\gamma$:
$$
k_{\gamma}(\boldx,\boldx') = \exp\left(- \frac{\|\boldx - \boldx'\|^2}{\gamma^2} \right).$$
A density-difference estimator $\diffh$ is obtained as
\begin{align*}
\diffh := \mathop{\arg \min }_{f \in \calHg}
\left[\|f\|_{\LRd}^2 - 2 Q_n f + \lambda \|f\|^2_{\calHg}\right].
\end{align*}

We assume the following conditions:
\begin{assumption}
\label{ass:AssOfTrueFunc}
The densities are bounded: There exists $M$ such that 
\begin{align*}
\|p\|_{\infty} \leq M
\ \ \mbox{and} \ \
\|p'\|_{\infty} \leq M.
\end{align*}
The density difference $\diff = p - p'$ is a member of Besov space with regularity $\alpha$:
$\diff \in B_{2,\infty}^{\alpha}$ and, for $r = \lfloor \alpha \rfloor + 1$
where $\lfloor \alpha \rfloor$ denotes the largest integer less than or equal to $\alpha$,
\begin{align*}
\|\diff \|_{B_{2,\infty}^{\alpha}} := \|\diff\|_{L_2(\Real^d)} + \sup_{t>0}(t^{-\alpha}\omega_{r,L_2(\Real^d)}(\diff,t)) < c,
\end{align*}
where $B_{2,\infty}^{\alpha}$ is the Besov space with regularity $\alpha$ and 
$\omega_{r,L_2(\Real^d)}$ is the $r$-th modulus of smoothness (see \emcite{NIPS2011_0874} for the definitions).
\end{assumption}

Then we have the following theorem.

\begin{theo}
Suppose Assumption \ref{ass:AssOfTrueFunc} is satisfied.
Then, for all $\epsilon>0$ and $p\in(0,1)$, there exists a constant $K>0$ depending on $M,c,\epsilon,p$ such that for all $n\geq 1$, $\tau \geq 1$, and $\lambda > 0$,
the LSDD estimator $\diffh$ in $\calHg$ satisfies
\begin{align*}
\|\diffh - \diff \|_{\LRd}^2 \!+\! \lambda \|\diffh\|_{\calHg}^2 \leq  K
\!\left( \lambda \gamma^{-d} \!+\! \gamma^{2\alpha} \!+\!
  \frac{\gamma^{-(1-p)(1+\epsilon)d}}{\lambda^p n} \!+\! 
\frac{\gamma^{-\frac{2(1-p)d}{1+p}(1+\epsilon + \frac{1-p}{4})}}{\lambda^{\frac{3p-p^2}{1+p}} n^{\frac{2}{1+p}}}
  \!+\! \frac{\tau}{n^2 \lambda}  \!+\! \frac{\tau}{n} \!\right)\!,
\end{align*}
with probability not less than $1- 4e^{-\tau}$.
\end{theo}

To prove this, we utilize the technique developed in \emcite{NIPS2011_0874}
for a regression problem.

\begin{proof}
First, note that 
$$
\|\diffh\|_{\LRd}^2 - 2 Q_n \diffh + \|\diff\|_{\LRd}^2 + \lambda \|\diffh\|_{\calHg}^2
\leq 
\|\fzero\|_{\LRd}^2 - 2 Q_n \fzero + \|\diff\|_{\LRd}^2 + \lambda \|\fzero\|_{\calHg}^2.
$$
Therefore, we have 
\begin{align}
&\!\!\!\!\!\!\!\!\!\!\!\!\!
\|\diffh - \diff \|_{\LRd}^2 + \lambda \|\diffh\|_{\calHg}^2 \notag \\
&= \|\diffh\|_{\LRd}^2 - 2Q_n \diffh + \|\diff\|_{\LRd}^2 + 2(Q_n - Q)\diffh + \lambda \|\diffh\|_{\calHg}^2 \notag\\
&\leq
 \|\fzero\|_{\LRd}^2 - 2Q_n \fzero + \|\diff\|_{\LRd}^2 + 2(Q_n - Q)\diffh + \lambda \|\diffh\|_{\calHg}^2 \notag\\
&= \|\fzero\|_{\LRd}^2 - 2Q \fzero + \|\diff\|_{\LRd}^2 + 2(Q_n - Q)(\diffh - \fzero) +  \lambda \|\diffh\|_{\calHg}^2 \notag\\
&= \|\fzero - \diff \|_{\LRd}^2 + 2(Q_n - Q)(\diffh - \diff) + 2(Q_n - Q)(\diff - \fzero) +  \lambda \|\diffh\|_{\calHg}^2.
\label{eq:BasicInequality}
\end{align}

Let 
$$
K(\boldx) := \sum_{j=1}^r {r \choose j} (-1)^{1-j} \frac{1}{j^d} \left(\frac{2}{\gamma \sqrt{\pi}} \right)^{\frac{d}{2}} 
\exp\left(- \frac{2\|\boldx\|^2}{j^2 \gamma^2} \right),
$$
and $\ftil(\boldx) := (\gamma \sqrt{\pi})^{-\frac{d}{2}} \diff$.
Using $K$ and $\ftil$, we define 
$$
\fzero := K * \ftil := \int_{\Real^d} \ftil(y)K(x-y) \mathrm{d}y,
$$
i.e., $\fzero$ is the convolution of $K$ and $\ftil$.
Because of Lemma 2 in \emcite{NIPS2011_0874}, we have $\fzero \in \calHg$ and 
\begin{align}
\|\fzero\|_{\calHg} & \leq (2^r -1) \|\ftil\|_{\LRd}~~(\because \text{Lemma 2 of \emcite{NIPS2011_0874}}) \notag \\ 
&\leq (2^r -1)(\gamma \sqrt{\pi})^{-\frac{d}{2}} \|\diff\|_{\LRd} \notag \\
&\leq (2^r -1)(\gamma \sqrt{\pi})^{-\frac{d}{2}} (\|p\|_{\LRd} + \|p'\|_{\LRd}) \notag \\
&\leq (2^r -1)(\gamma \sqrt{\pi})^{-\frac{d}{2}} 2 \sqrt{M}. 
\label{eq:fzeroHgbound}
\end{align}
Moreover, Lemma 3 in \emcite{NIPS2011_0874} gives
\begin{align}
\|\fzero \|_{\infty} \leq (2^r -1) \|\diff\|_{\infty} \leq (2^r -1) M,
\end{align}
and Lemma 1 in \emcite{NIPS2011_0874} yields that 
there exists a constant $C_{r,2}$ such that 
\begin{align}
\|\fzero - \diff \|_{\LRd}^2 \leq C_{r,2} \omega_{r,\LRd}^2(\diff,\frac{\gamma}{2}) \leq C_{r,2}c^2 \gamma^{2\alpha}.
\label{eq:fzeroL2bound}
\end{align}

Now, following a similar line to Theorem 3 in \emcite{NIPS2011_0874}, we can show that,
for all $\epsilon >0$ and $p\in(0,1)$, there exists a constant $C_{\epsilon,p}$ such that 
$$
|(P_n -P)(\diffh - \diff)| \leq \diffh - \diff.
$$

To bound this, we derive the tail probability of
$$(P_n - P)\left(\frac{\diffh - \diff}{\|\diffh - \diff\|_{\LRd}^2 + \lambda \|\diffh\|_{\calHg}^2+r}\right),$$
where $r>0$
is a positive real such that $r > r^*$ for
$$r^* = \min_{f\in \calHg} \|f - \diff\|_{\LRd}^2 + \lambda \|f\|_{\calHg}^2.$$
Let 
$$
g_{f,r} = \frac{f - \diff}{\|f - \diff\|_{\LRd}^2 + \lambda \|f\|_{\calHg}^2+r}
$$
for $f \in \calHg$ and $r > r^*$.
Then we have 
\begin{align*}
\|g_{f,r}\|_{\infty} 
& \leq \frac{\|f\|_{\infty} + \|\diff\|_{\infty}}{\|f - \diff\|_{\LRd}^2 + \lambda \|f\|_{\calHg}^2+r} \\
& \leq \frac{\|f\|_{\calHg} + \|\diff\|_{\infty}}{\|f - \diff\|_{\LRd}^2 + \lambda \|f\|_{\calHg}^2+r} \\
& \leq \frac{1}{\lambda \|f\|_{\calHg}+r/\|f\|_{\calHg}} + \frac{M}{r}  \leq \frac{1}{2\sqrt{r\lambda}} + \frac{M}{r},
\end{align*}
and
\begin{align*}
Pg_{f,r}^2 = \frac{P(f-\diff)^2}{(\|f - \diff\|_{\LRd}^2 + \lambda \|f\|_{\calHg}^2+r)^2}
\leq \frac{M\|f-\diff\|_{\LRd}^2}{(\|f - \diff\|_{\LRd}^2 + \lambda \|f\|_{\calHg}^2+r)^2}
\leq \frac{M}{r}.
\end{align*}
Here, let 
$$\calF_r:=\{f \in \calHg \mid \|f-\diff\|_{\LRd}^2 + \lambda \|f\|_{\calHg}^2 \leq r\},$$
and 
we assume that there exists a function such that
%that upper bounds $\mathbbE[\sup_{f \in \calF_r}|(P_n - P)(f-\diff)|]$:
$$
\mathbbE\left[\sup_{f \in \calF_r}|(P_n - P)(f-\diff)|\right] \leq \varphi_n(r),
$$
where $\mathbbE$ denotes the expectation over all samples.
Then, by the peeling device (see Theorem 7.7 in \npcite{Book:Steinwart:2008}), we have
$$
\mathbbE \sup_{f\in \calHg} |(P_n - P)g_{f,r}| \leq \frac{8\varphi(r)}{r}.
$$
Therefore, by Talagrand's concentration inequality, we have 
\begin{align}
\mathrm{Pr}\left[\sup_{f\in \calHg} |(P_n - P) g_{f,r}| < \frac{10 \varphi_n(r)}{r} + \sqrt{\frac{2M\tau}{nr}} + \frac{14 \tau}{3n}\left(\frac{1}{2\sqrt{r\lambda}} + \frac{M}{r}\right)\right]
\geq 1-e^{-\tau},
\end{align}
where $\mathrm{Pr}[\cdot]$ denotes the probability of an event.

From now on, we give an upper bound of $\varphi_n$.
The RKHS $\calHg$ can embedded in arbitrary Sobolev space $W^{m}(\Real^d)$.
Indeed, by the proof of Theorem 3.1 in \emcite{AS:Steinwart+Scovel:2004}, we have 
$$
\|f\|_{W^m(\Real^d)} \leq C_m \gamma^{-\frac{m}{2} + \frac{d}{4}} \|f\|_{\calHg}
$$
for all $f \in \calHg$.
Moreover, the theories of interpolation spaces give that, for all $f \in W^m(\Real^d)$, the supremum norm of $f$ can be bounded as 
$$
\|f\|_{\infty} \leq C_m' \|f\|_{\LRd}^{1-\frac{d}{2m}} \|f\|_{W^m(\Real^d)}^{\frac{d}{2m}},
$$
if $d < 2m$.
Here we set $m = \frac{d}{2p}$. Then we have 
$$
\|f\|_{\infty} \leq C_p'' \|f\|_{\LRd}^{1-p} \|f\|_{\calHg}^{p} \gamma^{-\frac{d(1-p)}{4}}.
$$
Now, since $\calF_r \subset (r/\lambda)^{1/2}\calB_{\calHg}$ and
$$P(f-\diff)^2 \leq M \|f-\diff\|_{\LRd}^2 \leq Mr~~~\mbox{for }f \in \calF_r$$
hold from Theorem 7.16 and Theorem 7.34 in \emcite{Book:Steinwart:2008}, we can take 
\begin{align*}
\varphi_n(r) = \max\Bigg\{
&C_{1,p,\epsilon} \gamma^{-\frac{(1-p)(1+\epsilon)d}{2}} \left(\frac{r}{\lambda}\right)^{\frac{p}{2}} (M r)^{\frac{1-p}{2}} n^{-1/2}, \\
&C_{2,p,\epsilon} \gamma^{-\frac{(1-p)(1+\epsilon)d}{1+p}} \left(\frac{r}{\lambda}\right)^{\frac{p}{1+p}} \left[\left(\frac{r}{\lambda}\right)^{\frac{p}{2}} \gamma^{-\frac{d(1-p)}{4}} r^{\frac{1-p}{2}}\right]^{\frac{1-p}{1+p}} n^{-1/(1+p)} \Bigg\},
\end{align*}
where $\epsilon>0$ and $p \in (0,1)$ are arbitrary and $C_{1,p,\epsilon},C_{2,p,\epsilon}$ are constants depending on $p,\epsilon$.
In the same way, we can also obtain a bound of $\sup_{f\in \calHg} |(P'_n - P') g_{f,r}|$.

If we set $r$ to satisfy
\begin{align}
\frac{1}{8} \geq \frac{10 \varphi_n(r)}{r} + \sqrt{\frac{2M\tau}{nr}} + \frac{14 \tau}{3n}\left(\frac{1}{2\sqrt{r\lambda}} + \frac{M}{r}\right),
\label{eq:TalagrandRHSbound}
\end{align}
then we have 
\begin{align}
\label{eq:QnQfhatftruediff}
|(Q_n - Q)(\diffh - \diff)| \leq \frac{1}{4}\left(r + \|\diffh - \diff\|_{\LRd}^2 + \lambda \|\diffh\|_{\calHg}\right),
\end{align}
with probability $1-2e^{-\tau}$.
To satisfy Eq.\eqref{eq:TalagrandRHSbound}, it suffices to set 
\begin{align}
r = C\left( \frac{\gamma^{-(1-p)(1+\epsilon)d}}{\lambda^p n} + \frac{\gamma^{-\frac{2(1-p)d}{1+p}(1+\epsilon + \frac{1-p}{4})}}{\lambda^{\frac{3p-p^2}{1+p}} n^{\frac{2}{1+p}}}
+
\frac{\tau}{n^2 \lambda} + \frac{\tau}{n}
\right),
\label{eq:rLargeBound}
\end{align}
where $C$ is a sufficiently large constant depending on $M,\epsilon,p$.

Finally, we bound the term $(Q_n - Q)(\fzero - \diff)$.
By Bernstein's inequality, we have 
\begin{align}
 |(P_n - P)(\fzero - \diff)| 
&\leq C \left( \|\diff - \fzero\|_{L_2(P)} \sqrt{\frac{\tau}{n}} + \frac{2^r M \tau}{n} \right)\notag \\
&\leq C \left( \sqrt{2M}\|\diff - \fzero\|_{\LRd} \sqrt{\frac{\tau}{n}}+ \frac{2^r M \tau}{n} \right)\notag \\
&\leq C \left( \|\diff - \fzero\|_{\LRd}^2 + \frac{2M\tau}{n} + \frac{2^r M \tau}{n}\right),
\end{align}
with probability $1-e^{-\tau}$,
where $C$ is a universal constant. In a similar way, we can also obtain 
$$
 |(P'_n - P')(\fzero - \diff)| \leq C \left( \|\diff - \fzero\|_{\LRd}^2 + \frac{2M\tau}{n} + \frac{2^r M \tau}{n}\right).
$$
Combining these inequalities, we have 
\begin{align}
\label{eq:QnQfzeroftruediff}
 |(Q_n - Q)(\fzero - \diff)| \leq C \left( \|\diff - \fzero\|_{\LRd}^2 + \frac{2^r M \tau}{n} \right),
\end{align}
with probability $1-2e^{-\tau}$, where $C$ is a universal constant.

Substituting Eqs.\eqref{eq:QnQfhatftruediff} and \eqref{eq:QnQfzeroftruediff} into Eq.\eqref{eq:BasicInequality}, 
we have 
\begin{align*}
&\|\diffh - \diff\|_{\LRd}^2 + \lambda \|\diffh\|_{\calHg}^2\\
&~~~\leq 
2\left\{ \|\fzero - \diff\|_{\LRd}^2 + 
C \left( \|\diff - \fzero\|_{\LRd}^2  + \frac{2^r M \tau}{n} \right)
+ r + \lambda \|\fzero\|_{\calHg} \right\},
\end{align*}
with probability $1-4e^{-\tau}$.
Moreover, by Eqs.\eqref{eq:fzeroL2bound} and \eqref{eq:fzeroHgbound}, the right-hand side is further bounded by 
\begin{align*}
\|\diffh - \diff\|_{\LRd}^2 + \lambda \|\diffh\|_{\calHg}^2
\leq 
C \left\{ \gamma^{2\alpha}  
+ r + \lambda \gamma^{-d} + \frac{1+\tau}{n} \right\},
\end{align*}
Finally, substituting \eqref{eq:rLargeBound} into the right-hand side,
we have 
\begin{align*}
&\|\diffh - \diff\|_{\LRd}^2 + \lambda \|\diffh\|_{\calHg}^2\\
&~~~\leq 
C \left\{ \gamma^{2\alpha}  
+ \frac{\gamma^{-(1-p)(1+\epsilon)d}}{\lambda^p n} + \frac{\gamma^{-\frac{2(1-p)d}{1+p}(1+\epsilon + \frac{1-p}{4})}}{\lambda^{\frac{3p-p^2}{1+p}} n^{\frac{2}{1+p}}} 
+ \lambda \gamma^{-d} + \frac{\tau}{\lambda n^2} +  \frac{\tau}{n} \right\},
\end{align*}
with probability $1-4e^{-\tau}$ for $\tau \geq 1$.
This gives the assertion.
\end{proof}

If we set 
$$
\lambda = n^{-\frac{2\alpha + d}{(2\alpha + d)(1+p) + (\epsilon -p + \epsilon p)}},~~
\gamma = n^{-\frac{1}{(2\alpha + d)(1+p) + (\epsilon -p + \epsilon p)}},
$$
and take $\epsilon, p$ sufficiently small, 
then we immediately have the following corollary.

\begin{Corollary}
\label{cor:TheMainBound}
Suppose Assumption \ref{ass:AssOfTrueFunc} is satisfied.
Then, for all $\rho,\rho'>0$, there exists a constant $K>0$ depending on $M,c,\rho,\rho'$ such that 
for all $n\geq 1$, $\tau \geq 1$, 
the density-difference estimator $\diffh$ with appropriate choice of $\gamma$ and $\lambda$
satisfies
\begin{align}
\|\diffh - \diff \|_{\LRd}^2 + \lambda \|\diffh\|_{\calHg}^2 \leq  
K \left( n^{-\frac{2\alpha}{2\alpha + d} + \rho} + \frac{\tau}{n^{1-\rho'}} \right),
\end{align}
with probability not less than $1- 4e^{-\tau}$.
\end{Corollary}

Note that $n^{-\frac{2\alpha}{2\alpha + d}}$ is the optimal learning rate to estimate a function in $B_{2,\infty}^\alpha$ \cite{NIPS2011_0874}.
Therefore, the density-difference estimator with a Gaussian kernel achieves the optimal learning rate
by appropriately choosing the regularization parameter and the Gaussian width.
Because the learning rate depends on $\alpha$,
the LSDD estimator has an adaptivity to the smoothness of the true function.

Our analysis heavily relies on the techniques developed in \emcite{NIPS2011_0874}
for a regression problem.
The main difference is that
the analysis in their paper involves a clipping procedure,
which stems from the fact that
the analyzed estimator
requires an empirical approximation of the expectation of the square term.
% $\EE[f^2]$.
The Lipschitz continuity of the square function $f \mapsto f^2$ 
is utilized to investigate this term,
and the clipping procedure is used to ensure the Lipschitz continuity.
On the other hand, in the current paper,
we can exactly compute $\|f\|_{\LRd}^2$ so that
we do not need the Lipschitz continuity.

\section{Derivation of Eq.\eqref{L2h-alpha-theta-hh--(1-alpha)-theta-H-theta-expansion}}
\label{proof-theo:L2h-alpha-theta-hh--(1-alpha)-theta-H-theta-expansion}
  When $\lambda$ ($\ge0$) is small,
  $\left(\boldH+\lambda\boldI_{\nparam}\right)^{-1}$ can be expanded as
  \begin{align*}
    \left(\boldH+\lambda\boldI_{\nparam}\right)^{-1}
    =\boldH^{-1}-\lambda\boldH^{-2}+o_p(\lambda),
  \end{align*}
  where $o_p$ denotes the probabilistic order.
  Then Eq.\eqref{L2h-alpha-theta-hh--(1-alpha)-theta-H-theta} can be expressed as
\begin{align*}
  &\!\!\!\!\!\!\!\!\!\!
  \beta\boldhh^\top\boldthetah+(1-\beta)\boldthetah^\top\boldH\boldthetah\nonumber\\
  &=
  \beta\boldhh^\top\left(\boldH+\lambda\boldI_{\nparam}\right)^{-1}\boldhh
  +(1-\beta)\boldhh^\top\left(\boldH+\lambda\boldI_{\nparam}\right)^{-1}\boldH
  \left(\boldH+\lambda\boldI_{\nparam}\right)^{-1}\boldhh \nonumber\\
  &=
  \beta\boldhh^\top\boldH^{-1}\boldhh
  -\lambda\beta\boldhh^\top\boldH^{-2}\boldhh\nonumber\\
  &\phantom{=}
  +(1-\beta)\boldhh^\top\boldH^{-1}\boldhh
  -2\lambda(1-\beta)\boldhh^\top\boldH^{-2}\boldhh
  +o_p(\lambda)\nonumber\\
  &=\boldhh^\top\boldH^{-1}\boldhh-\lambda(2-\beta)\boldhh^\top\boldH^{-2}\boldhh
  +o_p(\lambda),
%  \label{L2h-alpha-theta-hh--(1-alpha)-theta-H-theta-expansion}
\end{align*}
which concludes the proof.

\section{Derivation of Eq.\eqref{hh-Hinv-hh-bias}}
\label{proof-theo:hh-Hinv-hh-bias}
Because $\mathbbE[\boldhh]=\boldh$, we have
\begin{align*}  
\mathbbE[\boldhh^\top\boldH^{-1}\boldhh-\boldh^\top\boldH^{-1}\boldh]
&=\mathbbE[(\boldhh-\boldh)^\top\boldH^{-1}(\boldhh-\boldh)]\\
&=\tr{\boldH^{-1}\mathbbE[(\boldhh-\boldh)(\boldhh-\boldh)^\top]}\\
&=\tr{\boldH^{-1}\left(\frac{1}{\nsample}\boldV_{\density}[\boldpsi]
    +\frac{1}{\nsample'}\boldV_{\density'}[\boldpsi]\right)},
\end{align*}
which concludes the proof.

\end{document}

%% file: mathdef.tex
\newtheorem{defi}{Definition}

\newtheorem{theo}[defi]{Theorem}

\newtheorem{assumption}[defi]{Assumption}
\newcommand{\argmin}{\mathop{\mathrm{argmin\,}}}

\newcommand{\tr}[1]{\mathrm{tr}\left(#1\right)}

\newcommand{\iid}{\stackrel{\mathrm{i.i.d.}}{\sim}}

\newcommand{\mathbbE}{\mathbb{E}}

\newcommand{\mathbbR}{\mathbb{R}}

\newcommand{\boldzero}{{\boldsymbol{0}}}

\newcommand{\boldH}{{\boldsymbol{H}}}
\newcommand{\boldI}{{\boldsymbol{I}}}

\newcommand{\boldV}{{\boldsymbol{V}}}

\newcommand{\boldY}{{\boldsymbol{Y}}}

\newcommand{\boldc}{{\boldsymbol{c}}}

\newcommand{\boldh}{{\boldsymbol{h}}}

\newcommand{\boldx}{{\boldsymbol{x}}}
\newcommand{\boldy}{{\boldsymbol{y}}}

\newcommand{\boldtheta}{{\boldsymbol{\theta}}}

\newcommand{\boldmu}{{\boldsymbol{\mu}}}

\newcommand{\boldpsi}{{\boldsymbol{\psi}}}

\newcommand{\boldSigma}{{\boldsymbol{\Sigma}}}

\newcommand{\calB}{{\mathcal{B}}}

\newcommand{\calF}{{\mathcal{F}}}

\newcommand{\calX}{{\mathcal{X}}}
\newcommand{\calY}{{\mathcal{Y}}}

\newcommand{\boldthetah}{{\widehat{\boldtheta}}}

\newcommand{\inputdim}{d}

\newcommand{\diffsymbol}{f}
\newcommand{\diff}{\diffsymbol}
\newcommand{\diffh}{\widehat{\diffsymbol}}

\newcommand{\diffmodel}{g}

\newcommand{\densitysymbol}{p}

\newcommand{\density}{\densitysymbol}

\newcommand{\densityh}{\widehat{\densitysymbol}}

\newcommand{\nsample}{n}

\newcommand{\nparam}{b}

\newcommand{\calXt}{\widetilde{\calX}}

\newcommand{\boldhh}{{\widehat{\boldh}}}

%% file: mathdef_suzuki.tex
\newcommand{\dd}{\mathrm{d}}

\newcommand{\ftil}{\widetilde{f}}

\newcommand{\Real}{\mathbb{R}}

\newtheorem{Corollary}{Corollary}

\newcommand{\fzero}{f_0}

\newcommand{\calHg}{\mathcal{H}_{\gamma}}

\newcommand{\LRd}{L^2(\Real^d)}